\documentclass[lettersize,journal]{IEEEtran}
\usepackage[utf8]{inputenc}
\usepackage{amsmath,amsfonts}
\usepackage{algorithmic}
\usepackage{algorithm}
\usepackage{array}
\usepackage[caption=false,font=footnotesize]{subfig}
\usepackage{textcomp}
\usepackage{stfloats}
\usepackage{url}
\usepackage{verbatim}
\usepackage{graphicx}
\usepackage{cite}
\usepackage{tabularx}
\usepackage{tikz}
\usetikzlibrary{positioning,calc,fit,arrows,decorations.pathreplacing}
\tikzset{>=stealth}
\usepackage{siunitx, booktabs, multirow}

\begin{document}

\title{Geometry-Aware Surrogate for Real-Time Hydrodynamics Estimation of Autonomous Ground Vehicles in Amphibious Environments}

\author{Ammar Waheed, Luke Gallantree, and Zohaib Hasnain*
\thanks{Ammar Waheed and Zohaib Hasnain are with the J. Mike Walker ’66 Department of Mechanical Engineering, Texas A\&M University, College Station, TX 77801, USA.}
\thanks{Luke Gallantree was with the Mounted Systems Engineering Team, Platform Systems Division, Defence Science and Technology Laboratory, Portsdown West, PO17 6AD, UK.}
\thanks{This work has been supported by a grant from the Defence Science and Technology Laboratory (DSTL), United Kingdom.}
\thanks{*corresponding author: zhasnain@tamu.edu}
}


\maketitle

\begin{abstract} 
Autonomous ground vehicles operating in shallow water or flood-prone terrains require dynamic models that account for hydrodynamic forces. However, the simulation and planning tools currently available either lack the physical fidelity or are too computationally expensive to run in real time. This work presents a per-surface neural network surrogate that bridges this gap by predicting geometry-resolved hydrodynamic forces at real-time rates, trained entirely on high-fidelity CFD data from two geometrically distinct vehicles. A vehicle specific Signed Distance Field (SDF) provides per-surface submergence inputs, allowing the model to resolve how loading varies with vehicle geometry, depth, and flow direction. On held-out CFD data, the surrogate achieves a longitudinal-force symmetric MAPE (sMAPE) of 13\% and a vertical-force sMAPE of 3-12\%, with inference running under 0.9\,ms per sample. To evaluate the model under real-world conditions, water wading trials of a full-scale vehicle at different submersion depths are used. Motion capture derived kinematics serve as the surrogate inputs, and the resulting predictions are tested to reproduce known physical relationships between force, speed, and depth. The predicted drag follows quadratic speed scaling ($R^2 \geq 0.97$) and the buoyancy intercepts scale linearly with depth ($R^2 = 0.973$). Neither relationship is encoded in the model training loss, both emerge from the per-surface architecture summing individually predicted surface forces. The resulting framework provides a pathway for embedding physically grounded hydrodynamics into the simulation and planning loops that autonomous ground vehicles depend on in amphibious environments.
\end{abstract}

\renewcommand{\abstractname}{Note to Practitioners}
\begin{abstract} Autonomous vehicles are increasingly expected to operate in amphibious environments, where fluid forces significantly affect motion and safety. However, existing simulation and planning tools either ignore these effects or are too slow for practical use, making it difficult for engineers to design and test reliable systems. This paper presents a fast learned modeling approach that captures how water interacts with a vehicle as it moves. This enables engineers to include these effects in simulation and control design without requiring repeated expensive fluid simulations or extensive field testing. As a result, it can improve the reliability of vehicle behavior in challenging environments while reducing development time and cost. Experiments with a full-scale vehicle show that the model recovers established physical relationships under real operating conditions, even though these behaviors were not explicitly programmed. This suggests the approach can generalize well beyond the training data. The proposed framework supports a range of applications, including real-time controller tuning, motion planning, and mobility assessment in fluid-affected terrains.
\end{abstract}


\section{Introduction}
\IEEEPARstart{A}{ccurate} simulation is increasingly central to the development and testing of autonomous ground vehicles. For scenarios involving environmental complexity or safety-critical conditions, physical testing alone is often impractical due to high costs, limited repeatability, and risk to hardware \cite{pmlr-v78-dosovitskiy17a}. This is especially true in fluid-affected environments such as shallow water or flood-prone terrain. The ability to predict how a vehicle will behave in such settings requires simulation tools that capture not just surface-level appearance, but also the underlying physics governing vehicle-environment interaction. Prior work has shown that even advanced simulators can exhibit significant sim-to-real gaps when key physical effects are omitted or overly simplified \cite{choi2021use, s2r-survey}. 
\par
One critical yet under-represented interaction in many vehicle simulation frameworks is the dynamic influence of surrounding fluids on partially submerged systems. As ground vehicles enter shallow water or saturated terrain, hydrodynamic forces including drag, buoyancy, and added mass begin to dominate the dynamics \cite{GU2018343}. These forces are not merely environmental perturbations; they alter the effective inertia, weight distribution, and net resistive forces experienced by the vehicle \cite{UNav, matveev2025}. The interplay between vehicle geometry, velocity, submersion depth, and local flow state gives rise to highly nonlinear, coupled fluid-structure interactions that fundamentally change how the vehicle accelerates, turns, or maintains stability \cite{PAN2023114618}.
\par
Such interactions cannot be neglected in applications ranging from unmanned amphibious mobility to flood-zone navigation, where decisions made by autonomous agents rely on underlying dynamic models. If the simulator fails to capture effects such as the transient drag increase when a platform pitches forward into water \cite{huang2021cfd,kamath2017study} or the lift-induced reduction in tire normal force \cite{horne1963phenomena}, the resulting models will be incomplete. As a result, controllers or planners developed in such a simulator may perform poorly or fail when applied in real-world conditions.
\par
Traditional solutions for modeling these coupled dynamics rely on Computational Fluid Dynamics (CFD) solvers, which can resolve pressure distributions, flow separation, and time-dependent hydrodynamic loads with high fidelity \cite{varshney2022}. However, these methods are not feasible for interactive or real-time applications. Even simplified Navier-Stokes solvers with moving boundary conditions require extensive spatial discretizations and small time steps, often running several orders of magnitude slower than real-time, and full 3D simulations of vehicles transitioning between air, water, and ground modes can require hundreds of CPU-core hours to simulate seconds of physical time. 
\par
At the other extreme, many robotics simulators and game engines approximate fluid effects with fixed drag coefficients or simplified buoyancy primitives, achieving interactive frame rates but without computing loads from the underlying flow physics \cite{aldhaheri2025, de2022simu2vita, DAVE, Stonefish, song2025oceansim}. These models do not resolve how forces vary with vehicle geometry, submersion state, or flow direction, and consequently the dynamics they produce do not transfer reliably to real hardware. The result is a persistent gap; the tools that are physically accurate are too slow for real-time use, and the tools that are fast enough lack the fidelity needed for meaningful sim-to-real transfer.
\par
This work addresses the aforementioned gap through a data-driven surrogate model trained on high-fidelity CFD data that predicts spatially distributed hydrodynamic forces across the vehicle body at real-time rates. The surrogate operates at the level of individual surface regions rather than producing a single lumped force, allowing it to resolve how loading changes with geometry and submersion state. The model is trained on two geometrically distinct platforms and validated experimentally on a full-scale vehicle at multiple submersion depths using only the measured kinematics as input. The experimental results show that the surrogate reproduces established hydrodynamic scaling laws that are absent from the training objective. By making geometry-aware force estimates available at interactive rates, this approach enables the kind of physically grounded hydrodynamic modeling that simulation and planning pipelines for autonomous ground vehicles currently lack.
\section{Literature Review}
Understanding the behavior of vehicles in shallow water or flood‑like conditions requires modeling hydrodynamic forces that are absent on dry terrain. When a vehicle is partially submerged, buoyancy, drag, lift, and added mass become prominent and alter stability and mobility. Several studies focusing on amphibious and flood-driven vehicles have highlighted the importance of these forces for both mobility and stability \cite{more2014stability, xia2014criterion, hu2023experimental}. Experimental and numerical investigations have defined critical thresholds for "incipient motion", where vehicles begin to float or slide under modest water flow, leading to instability \cite{bocanegra2020review}. 
\par
High-fidelity studies in the vehicle hydrodynamics literature solve the incompressible Navier-Stokes equations with free-surface tracking to resolve pressure distributions and unsteady forces on vehicle bodies. RANS-based workflows have been used to quantify amphibious-vehicle resistance \cite{liu2023resistance}, and form drag has been shown to account for 40-80\% of total resistance at low planing speeds \cite{PAN2023114618}. Tightly coupled multibody-CFD simulations have further demonstrated the interdependence of terrain mechanics and coastal hydrodynamics during land-to-water transitions \cite{yamashita2024}. However, the fluid-structure coupling is nonlinear \cite{hou2012numerical} and depends on immersion level, speed, body geometry, and transient motions. The diversity of vehicle platforms, from compact wheeled robots to large tracked carriers, introduces further variation in wave patterns and added-mass effects \cite{newman1977marine}, so that no single analytical model covers all geometries and operating regimes. Consequently, accurate force prediction requires either high-fidelity simulation or calibration from empirical data.

\subsection{Computational Fluid Dynamics (CFD) and Fluid-Structure Interaction (FSI)}
The most rigorous approach to modeling hydrodynamic forces solves the
Navier-Stokes equations over a domain containing the vehicle and
surrounding fluid. When an air-water interface is present, methods
such as Volume-of-Fluid (VOF) \cite{hirt1981volume} are used to track
the free surface. Together, these tools resolve pressure
distributions, wake separation, and dynamic forces with high fidelity.
\par
However, fine spatial grids, small time steps, and iterative
fluid-multibody coupling make full CFD simulations expensive. A few
seconds of physical time can require hours of computation
\cite{yamashita2024}. Simplifications such as shallow-water equations
or potential-flow theory reduce cost but sacrifice viscous and
nonlinear effects. GPU-accelerated solvers (e.g.\ Project Chrono
\cite{tasora2015chrono}) improve throughput yet remain oriented toward
offline analysis rather than real-time control. High-fidelity CFD
therefore serves as an accuracy reference but is impractical for
robotics tasks that require rapid iteration or extended simulation.

\subsection{Particle-Based Methods}
Particle-based fluid solvers, especially Smoothed Particle Hydrodynamics (SPH), offer an alternative to grid-based CFD by representing the fluid as discrete particles that carry mass, pressure, and velocity. This representation is well suited for free-surface flows, complex boundary geometries, and multiphase contact.
\par
SPH has gained traction in robotics for simulating interactions with water. Angelidis et al.\ \cite{9982036} introduced an SPH-based plugin for the Gazebo simulator, enabling buoyancy and drag forces to be modeled through particle interactions for articulated robots. Project Chrono \cite{tasora2015chrono} similarly uses SPH to simulate two-way coupling between a fluid and multibody vehicle models. These particle-based frameworks can account for dynamic flow-induced torques and wave interactions that simpler models overlook. However, higher fidelity requires more particles, which slows execution: simulating a few seconds of physical time at high resolution can take hours, even with GPU acceleration.
\par
Hybrid methods such as Fluid-Implicit Particles (FLIP) reduce numerical dissipation by coupling particle tracking with a background grid. Macklin et al. \cite{macklin2014unified} implemented a position-based dynamics scheme that combines SPH and FLIP elements, achieving interactive frame rates on graphical hardware. Although visually compelling, such solvers prioritize stability over physical accuracy and lack real-time integration with control systems, making them unsuitable for force-level validation.
\subsection{Interactive Platforms and Simplified Force Models}
For applications requiring real-time execution, most game engines and robotic simulators use simplified fluid models that prioritize visual fidelity and computational efficiency over physical accuracy. The Unreal Engine~5 built-in buoyancy model, for instance, uses spherical approximations that do not resolve the underlying fluid physics \cite{epic_water_buoyancy_component}. Hydrodynamic drag is often approximated with heuristic damping forces rather than being derived from velocity and geometry.
\par
Similar abstractions exist in robotics tools. The UUV Simulator plugin for Gazebo \cite{manhaes2016uuvsim} implements Fossen's equations of motion \cite{fossen2021handbook} with constant added-mass and drag coefficients. This approach is reasonable for steady, fully submerged motion but fails to capture partial submersion, surface waves, or sudden water entry. Extensions such as Gazebo Fluids \cite{9982036} have been introduced to address these shortcomings, but adoption remains limited. Overall, conventional simulation platforms inadequately capture such nuanced interactions, resulting in performance mismatches between simulated and physical systems \cite{waheed2025quantifying, s2r-survey}. As a result, sim-to-real transfer suffers.

\subsection{Surrogate Modeling and Data-Driven Approaches}
Given the performance limitations of CFD and the inaccuracy of simplified models, there is growing interest in data-driven surrogates that approximate the fluid-structure interaction map at a fraction of the computational cost. Wang et al.\ \cite{wang2023prediction} trained deep neural networks on limited CFD data to predict external flow fields and hydrodynamic forces on cylindrical bodies, demonstrating that learned models can generalize across Reynolds numbers with modest training sets. Bai et al.\ \cite{bai2022data} applied a similar strategy to a manta-ray robot, using experimental hydrodynamic measurements to train a network that reproduced unsteady force profiles. Seyed-Ahmadi and Wachs \cite{seyed2022physics} proposed a physics-inspired neural network architecture for predicting forces and torques in particle-laden flows, embedding symmetry and scaling constraints directly into the network structure.
\par
In the vehicle mobility domain, Yamashita et al.\ \cite{yamashita2024} coupled an \textbf{LSTM} network with a multi-body dynamics solver to predict hydrodynamic loads on a vehicle traversing the land-water boundary. The LSTM captures temporal history but treats the vehicle as a single body and requires sequence-level training data, limiting its applicability to geometries and maneuvers not seen during training. More broadly, physics-informed neural networks (PINNs) \cite{raissi2019physics} have been shown to embed governing equations into the loss function, improving generalization in sparse-data regimes, though they have not yet been applied to real-time vehicle hydrodynamics.
\par
A common limitation across these approaches is that they model the vehicle as a monolithic body, predicting a single aggregate force vector. None decompose the loading onto individual geometric surfaces, which is necessary for capturing the distinct drag, buoyancy, and added-mass contributions that vary with local immersion. Furthermore, existing surrogates have been developed and validated on a single vehicle geometry, leaving open the question of whether a learned hydrodynamic mapping can generalize across platforms that differ in shape, scale, and surface topology without vehicle-specific retraining.

\section{Contribution to the state-of-the-art}
Real-time prediction of hydrodynamic forces on ground vehicles
remains an open problem, particularly because the loading is governed by a tightly coupled interaction between vehicle geometry, fluid conditions, and motion state. Existing methods are limited because they either solve the full fluid equations at computational costs that preclude real-time use, or resort to approximations that discard the dependence of forces on geometry and submersion depth. A more capable model is needed, one that retains the geometry-resolved fidelity of CFD while executing fast enough to serve as a force module inside a real-time simulation or planning loop.
\par
This work addresses that need through a per-surface neural network surrogate that predicts spatially distributed hydrodynamic forces across the vehicle body, trained entirely on high-fidelity CFD data from two geometrically distinct platforms. This approach goes beyond simply replacing CFD with a faster approximator. By decomposing forces at the surface level, the surrogate resolves how individual body regions contribute to the net load as a function of depth and flow, a distinction that lumped-parameter models fundamentally cannot make. The model is validated experimentally on a full-scale platform using motion capture measured kinematics as input, and the results show that it reproduces known hydrodynamic scaling laws that are absent from the training objective. With sub-millisecond inference on a single CPU core, the surrogate's computational overhead is well within the latency requirements of onboard control and real-time simulation pipelines. The model therefore establishes a sim-to-real transfer pathway for geometry-resolved hydrodynamic forces

\section{Methodology}
The proposed approach uses high-fidelity CFD simulations of two geometrically distinct vehicle platforms to generate per-surface force labels over a range of speeds, submersion depths, and fluid densities. A shared-weight neural network is trained on this data to predict spatially resolved hydrodynamic forces from a compact set of global and per-surface features. The trained surrogate is validated against real-world water wading experiments using physics-based scaling laws.

\subsection{Training Data Collection}
Training the surrogate requires high-fidelity CFD data capturing per-surface fluid loading on the vehicle. In a typical approach for partially submerged vehicles, a dynamic mesh with overset grids allows the body to move through a stationary fluid domain while the Navier-Stokes equations are solved concurrently for flow and motion. This provides high accuracy but at high computational cost due to mesh motion and interface updates. An alternative approach, common in fluid mechanics, fixes the rigid body in space and imposes a moving fluid field over the geometry, eliminating mesh motion while still resolving key flow structures. The present study adopts this fixed-body, moving-fluid strategy. 

\subsubsection{Coordinate System \& Geometry}
A right handed Cartesian coordinate system is used in the CFD simulations in this study. The \(Z\) axis is aligned with the nominal direction of flow over the vehicle, pointing from the rear of the vehicle towards the front. The \(Y\) axis points upward, normal to the ground plane. The \(X\) axis points to the right side of the vehicle when viewed from the rear. All distances, velocities and forces from CFD are reported in this global frame.
\par
Two different platforms are used to enable the surrogate model to learn geometry and scale dependent flow variations. The first platform, Clearpath Husky A200 is represented as a watertight rigid-body mesh located at the center of the computational domain. This domain has dimensions \(6 \times 5 \times 1.5\) m in the \(Z\), \(X\) and \(Y\) directions respectively, which provides sufficient distance upstream and downstream for the development of the free surface and wake structures. The second platform, Clearpath Warthog, is also modeled similarly in a separate domain sized \(10 \times 6 \times 3\) m in the \(Z\), \(X\) and \(Y\) directions respectively.

\subsubsection{Computational Domain}
A Volume of Fluid (VOF) formulation with two immiscible phases (air and the working fluid) represents the free surface. The VOF equation is solved implicitly with a compressive volume-fraction discretization to maintain a sharp interface. The vehicle geometry is embedded in a polyhedral volume mesh with three nested local refinement regions surrounding the vehicle to resolve the free surface, bow wave, and separated wake. Boundary-layer prism layers are placed on the ground plane to capture near-wall gradients in the open-channel flow.

\subsubsection{Solver Configuration}
A pressure-based transient solver is used with the \(k\text{-}\omega\) SST turbulence model, second-order spatial discretization for momentum, and a compressive scheme for the volume fraction. Inlet and outlet boundaries are configured as pressure boundaries with hydrostatic distributions computed from the specified free-surface level, forming an open-channel configuration with controlled depth. The remaining boundaries, including the ground plane, are no-slip walls. Each simulation runs for 4~s of physical time with fixed time steps of 0.02~s (Husky) and 0.005~s (Warthog), selected to maintain CFL~$\approx$~1.0, with a per-step residual convergence criterion of \(10^{-5}\). This yields a quasi-steady response of the free surface and force histories while still resolving the transient formation of waves and splash around the vehicles.
\begin{figure}
    \centering
    \includegraphics[width=1.0\linewidth]{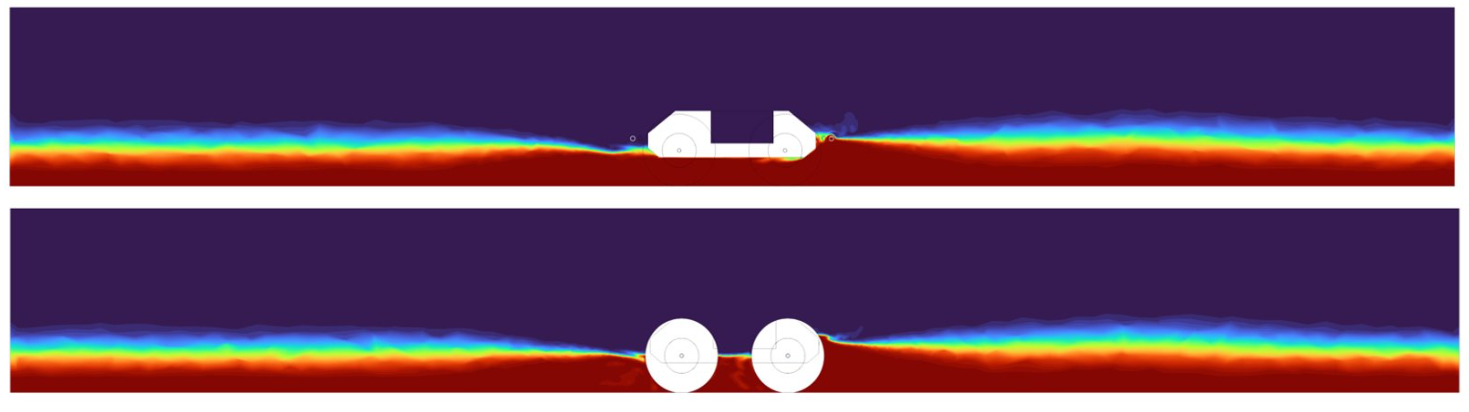}
    \caption{Volume fractions at two Z-Y planes; (top) center plane, (bottom) cross-sectional plane at wheels. Fluid flow direction is from right to left which represents vehicle moving towards right.}
    \label{fig:cfd}
\end{figure}

\subsubsection{Parameter Space}
The CFD setup described above is held fixed while a set of parameters is varied to generate a dataset suitable for surrogate model training. Table \ref{tab:cfd_params} summarizes the parameters, units, default ranges and increments used for the current study.
\begin{table}
\centering
\caption{Summary of parameters varied in the CFD data generation.}
\label{tab:cfd_params}
\begin{tabular}{lllll}
\hline
Sym. & Quantity & Units & Default range & Increment \\
\hline
\multicolumn{5}{c}{\textbf{Husky}} \\
\hline
\(U_{\infty}\) & Velocity magnitude & m/s & 0.2 to 1.0 & 0.2 \\
\(\phi\) & Flow angle (X-Z) & deg & 30 to 90 & 15 \\
\(\rho\) & Fluid density & kg/m\(^3\) & 1000 to 1900 & 300 \\
\(H\) & Water depth & m & 0.06 to 0.20 & 0.01 \\
\hline
\multicolumn{5}{c}{\textbf{Warthog}} \\
\hline
\(U_{\infty}\) & Velocity magnitude & m/s & 0.5 to 4.0 & 0.5 \\
\(\phi\) & Flow angle (X-Z) & deg & 30 to 90 & 15 \\
\(\rho\) & Fluid density & kg/m\(^3\) & 1000 to 1900 & 300 \\
\(H\) & Water depth & m & 0.10 to 0.40 & 0.05 \\
\hline
\end{tabular}
\end{table}
\par
The velocity ranges correspond to slow-to-moderate forward motion consistent with off-road operation. The flow angle \(\phi\) parametrizes the direction of the incoming flow in the horizontal plane, from head-on (\(\phi=90^\circ\)) to oblique (\(\phi=30^\circ\)), capturing a range of effective yaw angles. The fluid density spans fresh water (1000~kg/m\(^3\)) through heavy mud-like suspensions (1900~kg/m\(^3\)), and the depth range covers shallow splash conditions through deeper partial submergence relative to the wheel radius and chassis clearance.
\par
The parameter bounds in Table \ref{tab:cfd_params} define a multidimensional design space. The full tensor product grid would lead to a very large number of cases and is therefore not simulated in CFD in its entirety. Instead, Latin Hypercube Sampling (LHS) is applied within this space to select 175 parameter combinations that provide good coverage of the design space while keeping the total number of runs tractable.

\subsection{CFD Data Processing}
Fluent output files are parsed into a single structured format. Force time histories are extracted for 13 canonical surface regions on the Husky (3 front, 3 rear, 2 sides, underbody, 4 wheels) and 10 on the Warthog (1 front, 1 rear, 2 sides, 2 underbody, 4 wheels); the surface decomposition is discussed in detail in upcoming sections. All quantities are transformed from the Fluent coordinate system into a vehicle-centered frame in which \(X\) points forward, \(Y\) points left, and \(Z\) points upward, so that force components can be interpreted directly as drag, lateral force, and lift. Because the CFD uses a fixed body with a moving fluid, the velocity components are sign-inverted to represent vehicle velocity in still fluid. This conversion is applied before augmentation so that subsequent symmetry operations act on robot velocity.

\subsubsection{Transient Removal \& Section-based Averaging}
The consolidated time series for each case contains the entire 4 s simulation window with a fixed time step. The initial part of the simulation is dominated by the development of the flow and the free surface around the vehicle and is not representative of the statistically converged regime that is of interest for surrogate modeling. To remove this transient, only samples with physical time greater than 2 s are retained. 
\par
The remaining 2 s of data is then divided into a fixed number of contiguous sections of equal length. For the present work, 20 sections are used per case, each spanning 0.1 s of physical time. This count balances two competing concerns; too few sections over-smooth the temporal variability and reduce the effective training set size, while too many sections approach the raw time-step resolution and reintroduce transient fluctuations that are not representative of the quasi-steady regime. Each section-averaged sample represents the mean hydrodynamic response over its window, and the variability across sections within the same case arises from natural fluctuations in the resolved CFD flow field, including vortex shedding, free-surface oscillations, and wake unsteadiness, rather than from artificially injected noise. This natural variance acts as an implicit regularizer, preventing the surrogate from overfitting to a single time-averaged snapshot and encouraging it to learn a mapping that is robust to the stochastic component of the hydrodynamic load.

\subsubsection{Symmetry-based Augmentation}
The vehicle geometry exhibits bilateral and longitudinal symmetry, which is exploited in two successive augmentation steps. In the lateral augmentation, CFD cases are run only at positive flow angles (+30\(^\circ\) to +90\(^\circ\)); each sample is then mirrored about the vehicle's longitudinal symmetry plane by reflecting the lateral velocity, gravity, and force components and swapping left/right surface pairs. In the longitudinal augmentation, the forward velocity is sign-inverted to represent reverse motion, front and rear surface forces are swapped, and the streamwise force component is inverted. For surfaces on the symmetry plane (e.g., underbody), only the relevant directional component is inverted. Each step doubles the dataset, yielding a 4\(\times\) expansion overall.

\subsubsection{Augmented Dataset}
After augmentation, each sample consists of velocity components, scalar speed, fluid density, water depth, and gravity components as inputs, with three-component hydrodynamic forces on each surface as outputs. All quantities are in the vehicle-centered frame. This augmented dataset forms the basis for surrogate training.

\subsection{Geometric Definitions}
Before constructing the neural network dataset, the vehicle geometry is decomposed into \(\mathcal{X}\) semantically meaningful surface patches. The watertight Husky/Warthog mesh from the CFD setup is used to achieve this. The STL format meshes are loaded and a fixed axis transformation is applied. The resulting coordinate system is consistent with the CFD and SDF pipelines and defines a right-handed body frame in which the vehicle geometry is centered at the origin.
\par
The SDF computation and representation used in this study is identical for both Husky and Warthog. The Husky is divided into $K{=}13$ patches grouped by body region: three front panels (lower, mid, upper), three rear panels (lower, mid, upper), left and right side panels, a bottom panel, and four wheels (front-left, front-right, rear-left, rear-right). The Warthog uses 10 patches consisting of a single front panel, a single rear panel, two side panels, two underbody panels, and four wheels. The difference in patch count reflects geometric complexity. The Husky's tapered front and rear fairings change cross-sectional area rapidly with submersion depth, so lower, mid, and upper subdivisions are needed to resolve the depth-dependent transition in submerged area. The Warthog's flatter, more uniform front and rear panels are each adequately represented by a single patch, while its wider underbody is split into two panels to capture lateral variation in loading. The decomposition is therefore driven by how strongly each region's exposed area varies with submersion depth, not by the SDF grid resolution, which is uniform across the body. For simplicity, details for Husky are only mentioned further.
\par
For each patch, a set of aggregate geometric descriptors are computed. For a given surface name \(s\), the stored descriptor contains;
\begin{itemize}
    \item a semantic type string in \{\text{front}, \text{rear}, \text{side}, \text{bottom}, \text{wheel}\}
    \item a centroid position vector \(\mathbf{r}_s = (x_s, y_s, z_s)\) in meters
    \item an average unit normal vector \(\mathbf{n}_s\) obtained by averaging the mesh face normals for that patch and normalizing to unit length
    \item an area \(A_s\) representing the total surface area of all faces in the patch
\end{itemize}
\par
The patch normals follow the expected geometric orientations (lateral for sides, vertical for bottom, longitudinal for front/rear) and wheel patches carry an arbitrary normal since no single direction characterizes a circular shape.

\subsection{Signed Distance Field Representation}
A Signed Distance Field (SDF) is precomputed from the same STL mesh to support submergence calculations. An axis-aligned bounding box is computed around the transformed mesh, expanded by 0.5~m in all directions, and sampled on a regular 3D grid with spacing
\begin{equation}
\Delta
= \frac{\max\!\left(
\mathbf{b}_{\max} - \mathbf{b}_{\min}
\right)}{64}
\end{equation}
At every grid point the signed distance to the hull is evaluated using Python's \textit{Trimesh} proximity routines, yielding a scalar field that is negative inside the solid and positive outside. The field and grid metadata are stored for reuse across all dataset building steps.

\subsubsection{Depth Calculation from SDF}
The SDF and the patch labels are combined to precompute depth distributions for each surface. The dataset builder first loads the SDF volume and reconstructs the coordinate grids \((X,Y,Z)\) using the stored bounds and spacing, and computes the gradient of the SDF to obtain approximate hull normals throughout the grid. Although this gradient information is available for potential future features, the present work uses the SDF primarily as a convenient and highly precise volumetric representation of the hull geometry and relies on direct surface sampling for patch specific quantities.
\par
For each surface name \(s\) in the sorted list of labels, the corresponding face indices are used to extract a submesh of the vehicle geometry. The submesh inherits the same coordinate transform and scale as the SDF. A set of points is then sampled directly on this submesh using area proportionate sampling. From the sampled points associated with surface \(s\), the vertical coordinates \(z\) are extracted and filtered to remove points that fall below a fixed reference level that coincides with the ground plane used in the CFD simulations. The remaining vertical coordinates are stored as an array \(z_s\) which encodes the distribution of geometric heights for that surface relative to the Husky body frame. For each sampled point, the local surface normal from the submesh is also stored, yielding an array of normals \(\mathbf{n}_s^{(m)}\) indexed by sample.
\par
The result of this process is a geometry information structure that contains, for every surface patch, the set of vertical coordinates \(z_s\) and associated normals, together with the SDF grid metadata and the near hull mask. These patch level depth distributions are the basis for the submergence features used in the neural network input.

\subsubsection{Submergence Metrics}
Submergence features are computed for every sample by combining the per-patch depth distributions with the water depth. The water surface level is \(z_{\text{water}} = z_{0} + d\), where \(z_0\) is a fixed offset to the SDF coordinate system. A boolean mask identifies submerged sample points,
\begin{equation}
\chi_{i,s}^{(m)}
=
\begin{cases}
1, & z_{\text{water}}^{(i)} \ge z_{s}^{(m)}, \\
0, & \text{otherwise},
\end{cases}
\end{equation}
from which the submerged fraction and normalized mean depth are computed as
\begin{equation}
\text{sub\_frac}_{i,s}
= \frac{1}{M_s}
\sum_{m}
\chi_{i,s}^{(m)},
\end{equation}
\begin{equation}
\text{sub\_depth\_norm}_{i,s}
= \frac{1}{H_{\text{sub}}}
\frac{
\sum_{m}
\chi_{i,s}^{(m)}
\bigl(
z_{\text{water}}^{(i)} - z_{s}^{(m)}
\bigr)
}{
\sum_{m}
\chi_{i,s}^{(m)}
}
\end{equation}
where \(M_s\) is the number of sampled points on surface \(s\) and \(H_{\text{sub}} = 0.366\)~m is a characteristic submergence scale comparable to the Husky height. Completely dry surfaces receive zero for both metrics.

\section{Neural Network Architecture}
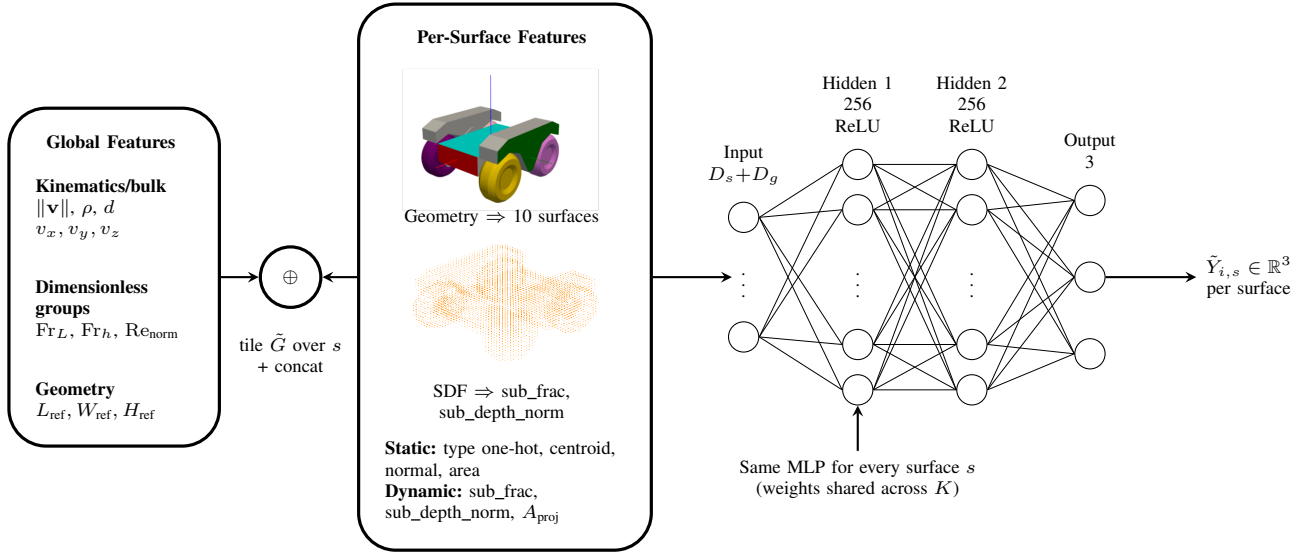
\begin{figure*}[t]
\centering
\begin{tikzpicture}[
  font=\scriptsize,
  panel/.style={draw, rounded corners=14pt, very thick, inner sep=10pt, align=center},
  paneltitle/.style={font=\bfseries, align=center},
  annot/.style={align=center},
  arr/.style={->, thick},
  conn/.style={draw, line width=0.30pt},
  neuron/.style={circle, draw, minimum size=4.0mm, inner sep=0pt},
]
\node[panel] (Gpanel) {
  \begin{minipage}[t]{0.11\textwidth}
    \centering
    {\textbf{Global Features}}\\[3mm]
    \raggedright
    \textbf{Kinematics/bulk}\\
    $\|\mathbf{v}\|$, $\rho$, $d$\\
    $v_x, v_y, v_z$\\[5mm]
    \textbf{Dimensionless groups}\\
    $\mathrm{Fr}_L$, $\mathrm{Fr}_h$, $\mathrm{Re}_{\text{norm}}$\\[5mm]
    \textbf{Geometry}\\
    $L_{\text{ref}}, W_{\text{ref}}, H_{\text{ref}}$\\[5mm]
  \end{minipage}
};

\node[panel, right=18mm of Gpanel] (Spanel) {
  \begin{minipage}[t]{0.17\textwidth}
    \centering
    {\textbf{Per-Surface Features}}\\[1mm]

    \includegraphics[width=1\linewidth]{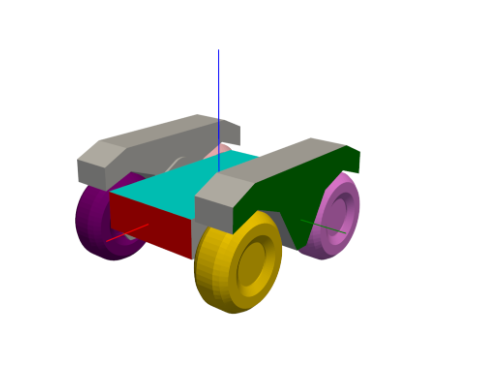}\\[-4mm]
    {\scriptsize Geometry $\Rightarrow$ 10 surfaces}\\[-5mm]

    \includegraphics[width=1\linewidth]{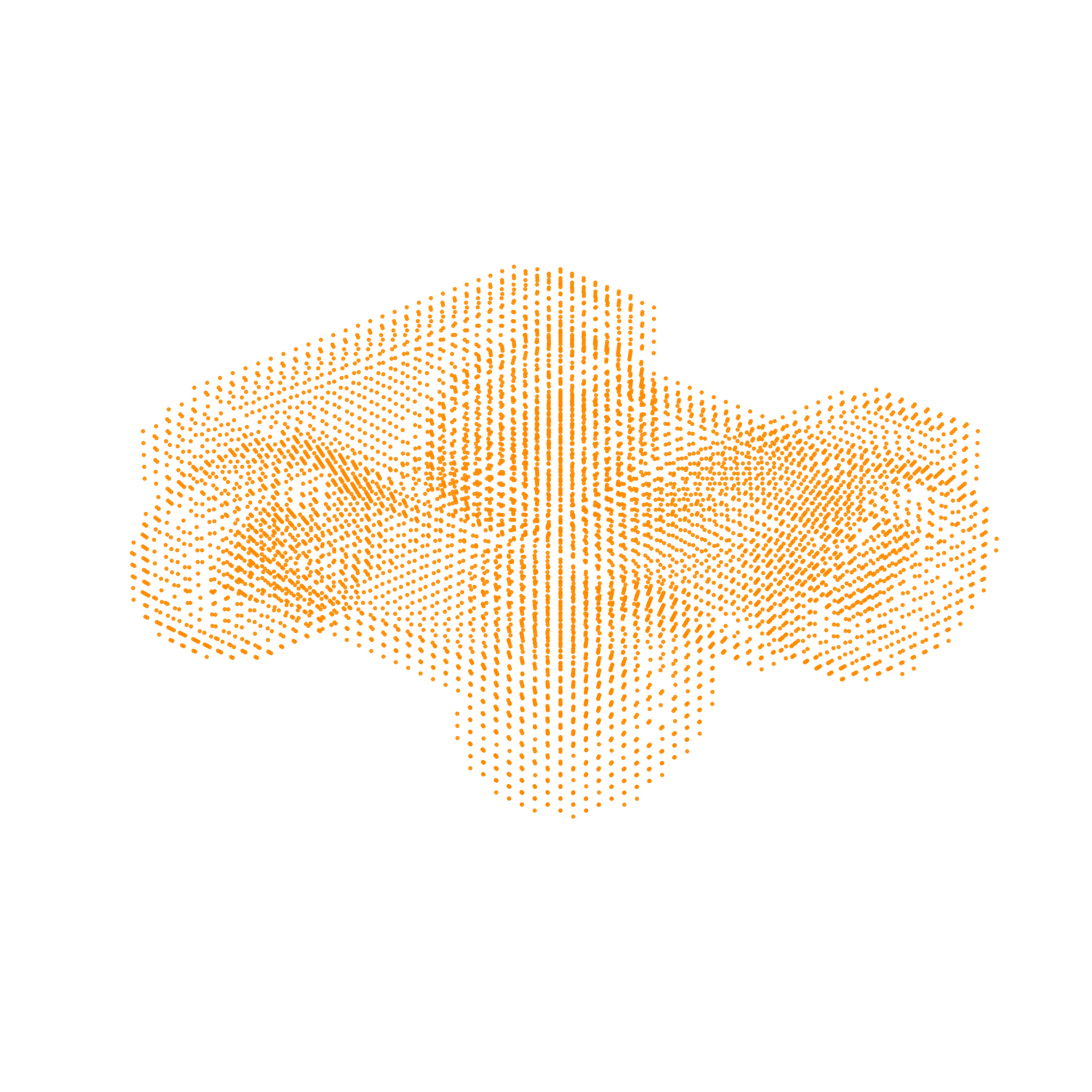}\\[-6mm]
    {\scriptsize SDF $\Rightarrow$ sub\_frac, sub\_depth\_norm}\\[2mm]

    \raggedright
    \textbf{Static:} type one-hot, centroid, normal, area\\
    \textbf{Dynamic:} sub\_frac, sub\_depth\_norm, $A_{\text{proj}}$
  \end{minipage}
};

\node[draw, circle, minimum size=8mm, very thick,
      right=0mm of Gpanel, xshift=5mm] (concat) {$\oplus$};

\draw[arr] (Gpanel.east) -- (concat.west);
\draw[arr] (Spanel.west) -- (concat.east);

\node[annot, below=2mm of concat] {\scriptsize tile $\tilde{G}$ over $s$ \\ \scriptsize + concat};


\coordinate (netStart) at ($(Spanel.east)+(12mm,0)$);

\node[annot] (Ic) at (netStart) {$\vdots$};
\node[neuron, above=2mm of Ic] (I1) {};
\node[neuron, below=2mm of Ic] (I2) {};
\node[annot, above=1mm of I1] (Ilabel) {Input\\$D_s{+}D_g$};

\node[annot, right=12mm of Ic] (H1c) {$\vdots$};
\node[neuron, above=9mm of H1c] (H11) {};
\node[neuron, above=3mm of H1c] (H12) {};
\node[neuron, below=3mm of H1c] (H13) {};
\node[neuron, below=9mm of H1c] (H14) {};
\node[annot, above=1mm of H11] (H1label) {Hidden 1\\256\\ReLU};

\node[annot, right=12mm of H1c] (H2c) {$\vdots$};
\node[neuron, above=9mm of H2c] (H21) {};
\node[neuron, above=3mm of H2c] (H22) {};
\node[neuron, below=3mm of H2c] (H23) {};
\node[neuron, below=9mm of H2c] (H24) {};
\node[annot, above=1mm of H21] (H2label) {Hidden 2\\256\\ReLU};

\node[neuron, right=12mm of H2c] (O2) {};
\node[neuron, above=6mm of O2] (O1) {};
\node[neuron, below=6mm of O2] (O3) {};
\node[annot, above=1mm of O1] (Olabel) {Output\\3};

\draw[arr] (Spanel.east) -- ++(10mm,0) |- (Ic.west);

\foreach \a in {I1,I2}{
  \foreach \b in {H11,H12,H13,H14}{
    \draw[conn] (\a.east) -- (\b.west);
  }
}
\foreach \a in {H11,H12,H13,H14}{
  \foreach \b in {H21,H22,H23,H24}{
    \draw[conn] (\a.east) -- (\b.west);
  }
}
\foreach \a in {H21,H22,H23,H24}{
  \foreach \b in {O1,O2,O3}{
    \draw[conn] (\a.east) -- (\b.west);
  }
}

\node[annot, right=12mm of O2] (Yshape) {$\tilde{Y}_{i,s}\in\mathbb{R}^{3}$\\per surface};
\draw[arr] (O2.east) -- (Yshape.west);

\node[annot, below=6mm of H14, xshift=0mm, text width=100mm] (sharednote)
{Same MLP for every surface $s$\\(weights shared across $K$)};
\draw[arr] (sharednote.north) -- (H14.south);
\end{tikzpicture}
\caption{Input-centric architecture of the hydrodynamic surrogate. Global features encode bulk flow state and dimensionless groups, while per-surface features encode patch type and geometry plus SDF-based submergence and projected exposure to the flow (Warthog shown as an example). By embedding these physical priors in the feature construction, the per-surface mapping can be learned with a small shared MLP, producing normalized force-per-density components for each surface.}
\label{fig:overview_inputs}
\end{figure*}
\subsection{Global Feature Matrix}
The global input features for each CFD sample are assembled into a matrix \(G \in \mathbb{R}^{N \times D_g}\). The base set of global features consists of;
\begin{equation}
\bigl\{
\lVert \mathbf{v} \rVert,\,
\rho,\,
d,\,
v_x,\,
v_y,\,
v_z
\bigr\}
\end{equation}

where, \((v_x, v_y, v_z)\) are the vehicle frame velocity components, \(|\mathbf{v}|\) is the scalar speed , \(\rho\) the fluid density, and \(d\) is the average submerged depth. These encode the kinematic and bulk flow state. On top of these base features, a set of dimensionless groups is appended. These include a body length Froude number;
\begin{equation}
\mathrm{Fr}_L
= \frac{\lVert \mathbf{v} \rVert}{\sqrt{g_{\text{eff}}\, L_{\text{ref}}}}
\end{equation}

and, a depth-based Froude number;
\begin{equation}
\mathrm{Fr}_h
= \frac{\lVert \mathbf{v} \rVert}{\sqrt{g_{\text{eff}}\, d}}
\end{equation}

which is clipped to a finite range to avoid unbounded values for very shallow depths, and a Reynolds number scaled to order one,
\begin{equation}
\mathrm{Re}_{\text{norm}}
= \frac{\rho\,\lVert \mathbf{v} \rVert\, L_{\text{ref}}}{\mu}
\times 10^{-6}
\end{equation}

where \(L_{\text{ref}}\) is the characteristic vehicle length, \(\mu = 1.002 \times 10^{-3}\) Pa s is the dynamic viscosity of water at \(20^\circ\)C, and \(g_{\text{eff}}\) is taken from the vertical component of the gravity vector as \(g = 9.81\) m s\(^{-2}\). Finally, the global feature vector includes normalized vehicle dimensions \((L_{\text{ref}}, W_{\text{ref}}, H_{\text{ref}})\). The full feature matrix \(G\) and the corresponding feature name list are returned by the dataset builder for later use in neural network training. 

\subsection{Per Surface Feature Tensor}
Local geometric and submergence information for each surface is encoded in a 3D tensor \(S \in \mathbb{R}^{N \times K \times D_s}\), where \(K\) is the number of surfaces. The feature set for each surface and sample is constructed as a concatenation of static and dynamic components.
\par
The static part depends only on the surface patch and is independent of the sample index. For each surface \(s\), a one-hot encoding of the semantic type is created using a fixed unified vocabulary (bottom, front, rear, side, wheel). The centroid position \(\mathbf{r}_s\) is normalized by the reference dimensions, yielding \((x_s / L_{\text{ref}}, y_s / W_{\text{ref}}, z_s / H_{\text{ref}})\). The average patch normal \(\mathbf{n}_s\) from the label file is appended directly, as it encodes the orientation of the patch in the body frame. The last static feature is a normalized area \(A_s / (L_{\text{ref}} H_{\text{ref}})\). These static features are precomputed once for all surfaces and then broadcast along the sample dimension. 
\par
The dynamic part depends on both the sample and the surface. It includes the submerged fraction \(\text{sub\_frac}_{i,s}\) and the normalized mean submerged depth \(\text{sub\_depth\_norm}_{i,s}\) defined earlier, and an orientation dependent projected area,
\begin{equation}
A_{\text{proj}, i,s}
= A_s\,\bigl|\mathbf{n}_s \cdot \hat{\mathbf{v}}^{(i)}\bigr|
\end{equation}
where \(\hat{\mathbf{v}}^{(i)}\) is the unit velocity vector for sample \(i\). This quantity measures the effective frontal area of the surface relative to the incoming flow direction and is normalized by \(L_{\text{ref}} H_{\text{ref}}\) before being stored as a feature. 
Together, these dynamic features allow the network to condition its predictions on both the instantaneous level of submergence and the orientation of each patch with respect to the flow.
\par
The full per surface feature tensor \(S\) is therefore composed of a static block that captures semantic type, location, orientation and size of the surface, and a dynamic block that captures submergence state and projected exposure to the flow.

\subsection{Force Targets \& Density Normalization}
The learning targets are assembled into a tensor \(Y \in \mathbb{R}^{N \times K \times 3}\). For each run, the CFD post processing provides the hydrodynamic force components \((F_{x,s}, F_{y,s}, F_{z,s})\) for all surfaces in the vehicle centered frame. To promote generalization across fluids with different densities, these forces are normalized by the corresponding density for each sample,

\begin{equation}
Y_{i,s,c}
= \frac{F_{c,s}^{(i)}}{\rho^{(i)}},
\qquad c \in \{x,y,z\}
\end{equation}

This representation encourages the neural network to learn a density independent mapping, so that scaling to new fluids such as salt water or mud mixtures can be handled by reapplying the appropriate density at inference time.

\subsection{Dataset Construction \& Input Tensors}
The consolidated dataset is converted into a PyTorch-compatible format. For each split, the following tensors are extracted:
\begin{itemize}
    \item global feature matrix \(G \in \mathbb{R}^{N \times D_g}\),
    \item per surface feature tensor \(S \in \mathbb{R}^{N \times K \times D_s}\),
    \item per surface target tensor \(Y \in \mathbb{R}^{N \times K \times 3}\),
    \item net force vector \(\mathbf{F}_{\text{net,true}} \in \mathbb{R}^{N \times 3}\) given by the sum of CFD forces over all surfaces.
\end{itemize}

During initialization, the dataset applies feature normalization based on statistics computed only from the training split. The normalization statistics are;
\begin{equation}
\boldsymbol{\mu}_G = \frac{1}{N_{\text{train}}} \sum_{i} G_i, \quad
\boldsymbol{\sigma}_G = \sqrt{\frac{1}{N_{\text{train}}} \sum_i (G_i - \boldsymbol{\mu}_G)^2},
\end{equation}

\begin{equation}
\boldsymbol{\mu}_Y = \frac{1}{N_{\text{train}} K} \sum_{i,s} Y_{i,s}, \quad
\boldsymbol{\sigma}_Y = \sqrt{\frac{1}{N_{\text{train}} K} \sum_{i,s} (Y_{i,s} - \boldsymbol{\mu}_Y)^2}
\end{equation}

Here \(Y_{i,s}\) denotes the three component target vector for sample \(i\) and surface \(s\). The normalized inputs and targets used for learning are;
\begin{equation}
\begin{aligned}
\tilde{G}_i
&= \frac{G_i - \boldsymbol{\mu}_G}{\boldsymbol{\sigma}_G}, \qquad
\tilde{Y}_{i,s}
= \frac{Y_{i,s} - \boldsymbol{\mu}_Y}{\boldsymbol{\sigma}_Y}
\end{aligned}
\end{equation}

while the surface features \(S\) and net forces \(\mathbf{F}_{\text{net,true}}\) are stored in physical units without normalization. Each call to the dataset returns a tuple \((\tilde{G}, S, \tilde{Y}, \mathbf{F}_{\text{net,true}}, sub_{frac})\) for a single sample. These tuples are batched by PyTorch into training and validation batches using a 80-20 split. The effective tensor shapes during training are summarized in Table \ref{tab:tensor_shapes}.
\begin{table}
\centering
\caption{Tensor shapes used in the training pipeline.}
\label{tab:tensor_shapes}
\begin{tabular}{llll}
\hline
Symbol & Description & Shape (batch \(B\)) & Units \\
\hline
\(\tilde{G}\) & norm. global features & \(B \times D_g\) & dimensionless \\
\(S\)         & Per surface features       & \(B \times K \times D_s\) & mixed \\
\(\tilde{Y}\) & norm. targets         & \(B \times K \times 3\) & dimensionless \\
\(\mathbf{F}_{\text{net,true}}\) & Net CFD force & \(B \times 3\) & N  \\
\(\text{sub\_frac}\) & Per surface wet frac. & \(B \times K\) & dimensionless \\
\hline
\end{tabular}
\end{table}

\subsection{Model Architecture}
The hydrodynamic surrogate is implemented as a surface-based Multi-Layer Perceptron (MLP). For a batch of samples, the network ingests the global features and per surface features and predicts the three components of force for each surface. The architecture can be written as;
\begin{equation}
\mathbf{z}_{i,s}
= \big[\, S_{i,s}, \tilde{G}_i \,\big]
\in \mathbb{R}^{D_s + D_g}
\end{equation}
\begin{equation}
\begin{aligned}
\mathbf{h}_1
&= \phi\!\left( W_1 \mathbf{z}_{i,s} + \mathbf{b}_1 \right), \\
\mathbf{h}_2
&= \phi\!\left( W_2 \mathbf{h}_1 + \mathbf{b}_2 \right)
\end{aligned}
\end{equation}
\begin{equation}
\tilde{Y}_{i,s}
= W_3 \mathbf{h}_2 + \mathbf{b}_3
\end{equation}

where \(\phi(\cdot)\) is the Rectified Linear Unit and \(W_1, W_2, W_3\) and \(\mathbf{b}_1, \mathbf{b}_2, \mathbf{b}_3\) are trainable weight matrices and bias vectors. The current implementation uses a hidden width of 256 units in both hidden layers and an output dimension of 3 corresponding to the three components of the normalized target vector per surface. The network constructs a batch tensor by tiling the global features along the surface dimension and concatenating them with the per surface features before passing them through a three layer fully connected MLP with ReLU activations. The output tensor has shape \(B \times K \times 3\) and represents the normalized force per unit density on each surface; because the same MLP is applied independently to each surface, the architecture naturally supports different \(K\) values across vehicles as long as mini-batches remain vehicle-homogeneous.

\subsection{Loss Function and Optimization}
The training objective is a single composite loss that combines (i) a per-surface hybrid loss in normalized space, (ii) a net-force consistency loss in density-normalized physical space, and (iii) a physics-violation penalty. The full loss is;
\begin{equation}
\mathcal{L}
= \mathcal{L}_{\text{hybrid}}
+ \lambda_{\text{net}}\,\mathcal{L}_{\text{net}}
+ \lambda_{\text{phys}}\,\mathcal{L}_{\text{phys}}
\label{eq:loss}
\end{equation}

The hybrid term is applied element-wise to the normalized tensor \(\tilde{Y}\) and is defined as
\begin{equation}
\mathcal{L}_{\text{hybrid}}
= (1 - \alpha)\,\mathcal{L}_{\text{MSE}}
+ \alpha\,\mathcal{L}_{\text{rel}}
\end{equation}
where
\begin{equation}
\mathcal{L}_{\text{MSE}}
= \frac{1}{3 B K}
\sum_{i,s,c}
\left(
\tilde{Y}_{i,s,c}^{\text{pred}}
- \tilde{Y}_{i,s,c}^{\text{true}}
\right)^{2},
\end{equation}
and
\begin{equation}
\mathcal{L}_{\text{rel}}
= \frac{1}{3 B K}
\sum_{i,s,c}
\left(
\frac{
\tilde{Y}_{i,s,c}^{\text{pred}}
- \tilde{Y}_{i,s,c}^{\text{true}}
}{
\max\!\left(
\left|\tilde{Y}_{i,s,c}^{\text{true}}\right|,
\varepsilon
\right)
}
\right)^{2}
\end{equation}
Here \(c \in \{x,y,z\}\) indexes the force components and \(\varepsilon = 0.01\) is a small positive constant that prevents division by very small target values. The mixing parameter is set to \(\alpha = 0.5\), which gives equal weight to the absolute and relative components in the normalized space. 
\par
To compute the net-force and physics penalty terms, predictions are first un-normalized back to density-normalized force units \(F/\rho\);
\begin{equation}
\hat{Y}_{i,s}
= \tilde{Y}_{i,s}^{\text{pred}} \odot \boldsymbol{\sigma}_Y
+ \boldsymbol{\mu}_Y, \qquad
Y_{i,s}
= \tilde{Y}_{i,s}^{\text{true}} \odot \boldsymbol{\sigma}_Y
+ \boldsymbol{\mu}_Y
\end{equation}

The net-force consistency loss compares summed per-surface forces (still in \(F/\rho\) units);
\begin{equation}
\mathcal{L}_{\text{net}}
= \frac{1}{3B}\sum_i
\left\|
\left(\sum_s \hat{Y}_{i,s}\right)
-
\left(\sum_s Y_{i,s}\right)
\right\|_2^2
\end{equation}

\(\mathcal{L}_{\text{phys}}\) penalizes non-zero predicted force on dry surfaces (\(\text{sub\_frac}<0.01\)). The final loss function is therefore computed as shown in Eq. \ref{eq:loss}. The hyperparameters listed in Table~\ref{tab:train_hparams} were selected through a coarse grid search on the validation loss, with the loss weights tuned to balance per-surface accuracy against net-force consistency. The optimization uses the Adam algorithm with a learning rate of \(10^{-3}\) and default momentum parameters \(\beta_1 = 0.9\) and \(\beta_2 = 0.999\). A learning rate scheduler monitors the validation loss and reduces the learning rate by a factor of 0.5 when the validation loss does not improve for 20 consecutive epochs.
\begin{table}
\centering
\caption{Hyperparameters used for training}
\label{tab:train_hparams}
\begin{tabular}{ll}
\hline
Quantity & Value \\
\hline
Training batch size & 16 \\
Validation batch size & 64 \\
Hidden layer width & 256 \\
Learning rate & \(10^{-3}\) \\
Hybrid loss weight \(\alpha\) & 0.5 \\
Relative error floor \(\varepsilon\) & 0.01 \\
Net consistency weight \(\lambda_{\text{net}}\) & 0.1 \\
Physics penalty weight \(\lambda_{\text{phys}}\) & 0.5 \\
Net force loss weight \(\lambda_F\) & 0.0 \\
Epochs & 1000 \\
\hline
\end{tabular}
\end{table}

\subsection{Validation Metrics}
After each epoch, the network is evaluated on the validation subset by computing a mean absolute error on the density-normalized net force (obtained by summing predicted \(F/\rho\) over surfaces);
\begin{equation}
\begin{aligned}
\text{MAE}_{\mathrm{net,norm}}
&= \frac{1}{3N_{\mathrm{val}}}\sum_{i}\sum_{c\in\{x,y,z\}}
\Biggl|
\sum_s \hat{Y}_{i,s,c}-\sum_s Y_{i,s,c}
\Biggr|
\end{aligned}
\end{equation}

Three ablation variants are also trained under the same configurations to evaluate the contribution of explicit vehicle dimensions, surface tessellation granularity, and the per-surface architecture (Sec.~\ref{sec:res_ablation}).

\section{Experimental Validation}
To evaluate the trained surrogate under real-world conditions, the Warthog platform was driven through a water channel at varied depths and speeds. The surrogate requires body-frame velocity and water-surface level as inputs, both of which must be derived from high-accuracy position measurements. Since onboard IMU and encoder data are insufficiently precise after differentiation, an outdoor-capable motion capture system with active infrared markers was therefore used to obtain millimeter-level position accuracy suitable for reliable velocity extraction.

\subsection{Experimental Setup}
\begin{figure}
    \centering
    \includegraphics[width=1\linewidth]{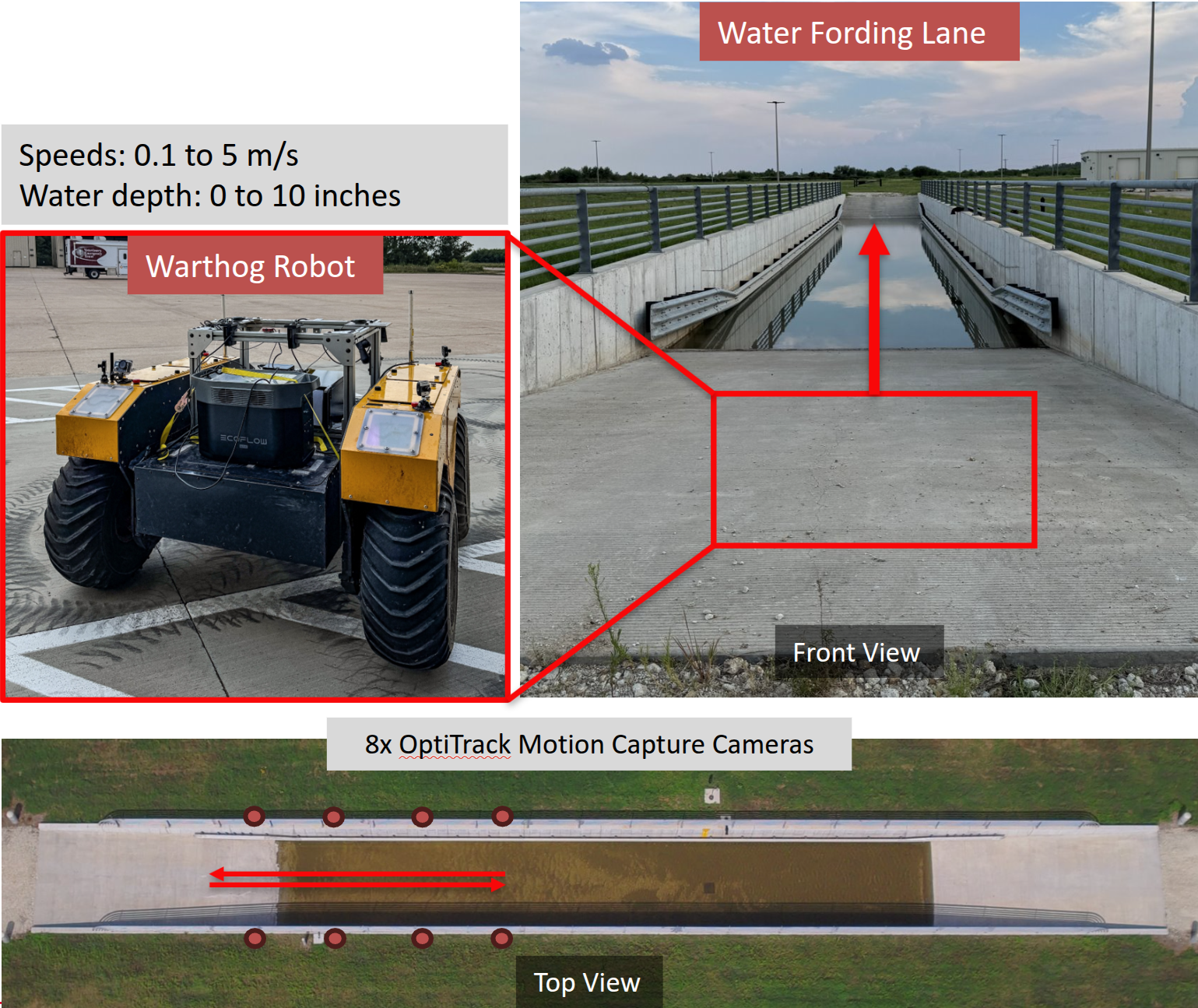}
    \caption{Experimental setup for water wading tests.}
    \label{fig:IPG}
\end{figure}

\begin{figure*}[t]
    \centering
    \subfloat[]{
        \includegraphics[width=0.32\textwidth]{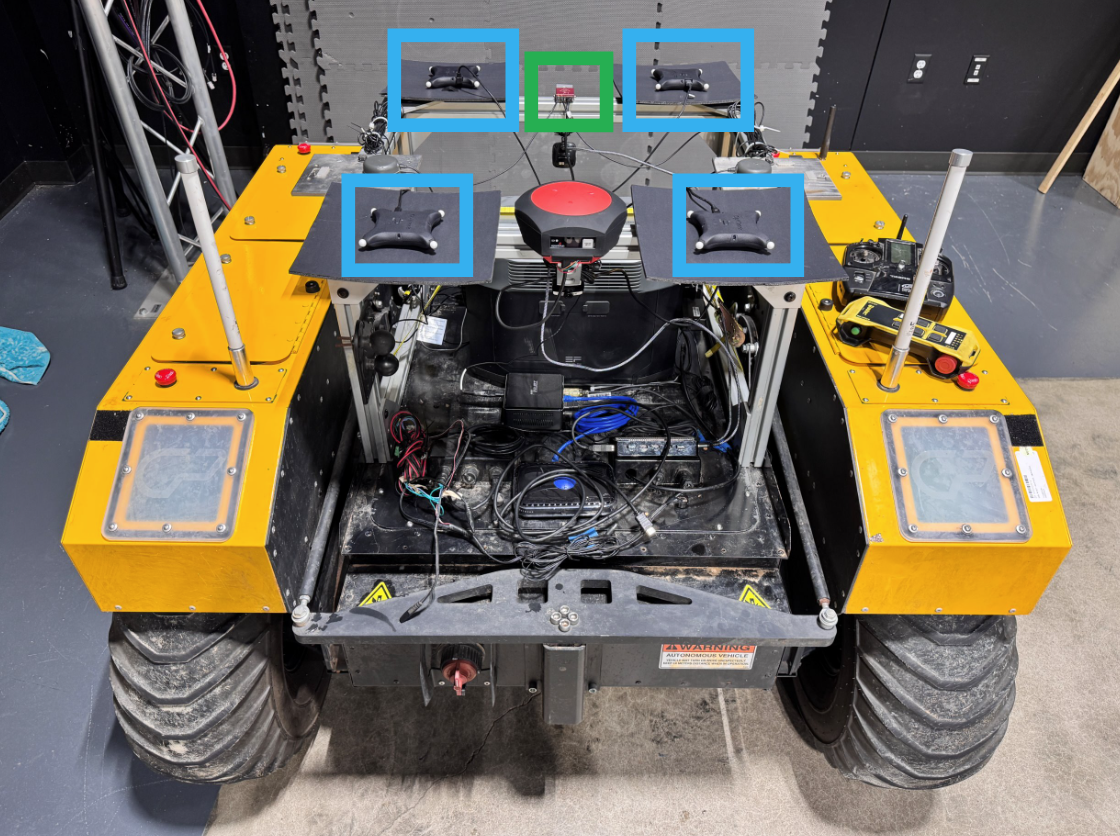}
        \label{fig:warthog_pucks}
    }
    \subfloat[]{
        \includegraphics[width=0.32\textwidth]{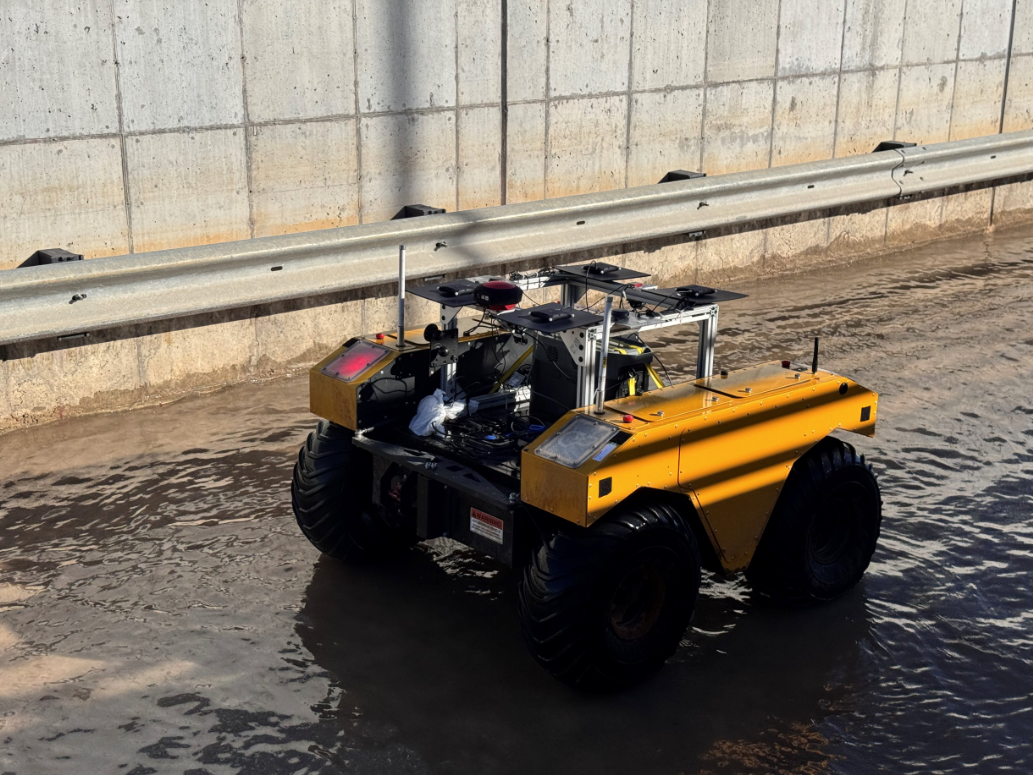}
        \label{fig:warthog_channel}
    }
    \subfloat[]{
        \includegraphics[width=0.32\textwidth]{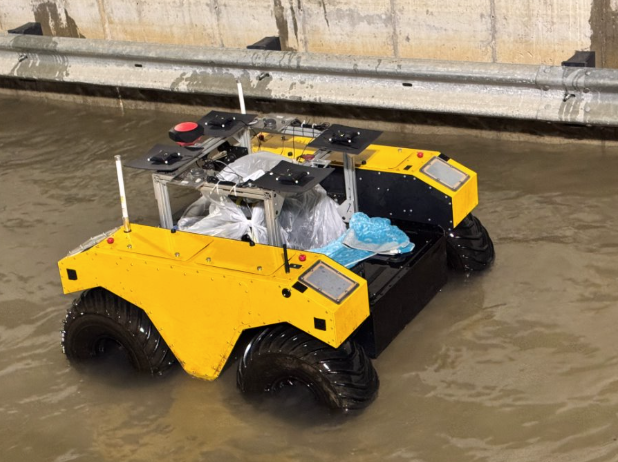}
        \label{fig:warthog_third}
    }
    \caption{Experimental setup; (a) shows the OptiTrack active pucks mounted onto the chassis frame, (b) shows the experiment run with 4 inches of water depth, (c) shows the experiment run with 10 inches of water depth}
    \label{fig:warthog}
\end{figure*}

Experiments were conducted at the Innovation Proving Grounds (IPG), Texas A\&M RELLIS campus. The facility includes a water fording lane, a channel approximately 60~m long, 6~m wide, and 2~m deep with ramp structures at each end (Fig.~\ref{fig:IPG}). The Clearpath Warthog, a skid-steer differential-drive platform capable of land and shallow-water operation, served as the test vehicle. Four OptiTrack active pucks were mounted on a rigid frame bolted directly to the chassis (Fig.~\ref{fig:warthog_pucks}), providing the rigid-body tracking needed for force-level motion analysis.

Eight OptiTrack PrimeX22 cameras were mounted along the water fording lane (four per side, Fig.~\ref{fig:IPG}) and configured to record at 120~Hz. Rigid-body pose (position and quaternion) was streamed via the Motive software and ingested into ROS using the \textit{mocap\_optitrack} package, synchronizing motion capture measurements with onboard data streams. Calibration yielded a maximum 3D marker error of 2.447~mm. The active pucks generate their own infrared illumination, enabling robust outdoor tracking independent of ambient lighting.

\subsection{Data Collection}
Test cases were organized around three primary control variables. The first variable was water depth in the channel, with target depths of 4, 8 and 10 inches, measured under quiescent conditions after each fill. Rest periods between successive trials allowed surface disturbances to dissipate before the next run. The second variable was the commanded forward velocity of the Warthog, which ranged between 0.5 m/s and 5 m/s, depending on depth and direction of travel. The third variable was the direction of motion relative to the ramps. For each depth, runs were collected for motion down the entry ramp into the channel and up the exit ramp out of the channel. Additional trials were conducted for an angled ramp in configuration at a depth of four inches. Table \ref{tab:test_cases} summarizes the 71 experiments recorded with the maximum and the mean commanded velocity per configuration.
\begin{table}[t]
\centering
\caption{Summary of Warthog water fording test cases.}
\label{tab:test_cases}
\scriptsize
\begin{tabularx}{\columnwidth}{lccXX}
\hline
Direction & Depth (in) & Cases & Max. Command Velocity (m/s) & Mean Command Velocity (m/s) \\
\hline
Ramp-in       & 4  & 18 & 4.952 & 2.984 \\
Ramp-out      & 4  & 17 & 4.952 & 2.708 \\
Ramp-in       & 8  & 9  & 3.025 & 1.978 \\
Ramp-out      & 8  & 5  & 2.730 & 1.654 \\
Ramp-in       & 10 & 9  & 2.531 & 1.616 \\
Ramp-out      & 10 & 9  & 2.284 & 1.550 \\
Ramp-in angle & 4  & 4  & 3.518 & 2.631 \\
\hline
\end{tabularx}
\end{table}

\subsection{Data Processing}
The motion capture system recorded the Warthog pose at 120 Hz and streamed it over the ROS network. Even with an average rate of 120 Hz, network transmission introduces small variations in the recorded timestamps, which produces a non-uniform sampling in time. For numerical differentiation, it is preferable to work with evenly spaced samples, so the raw motion capture data were resampled to a slightly lower but uniform rate of 111 Hz. Quaternion orientations were also interpolated to the new time grid and then re-normalized to preserve unit length. Position and quaternion histories were then smoothed and differentiated using a Savitzky-Golay filter (window length~9, polynomial order~2) to obtain translational velocity and angular velocity \(\boldsymbol{\omega}\). Because the marker origin is offset from the vehicle's Center of Mass (COM), the translational velocity is corrected via;
\begin{equation}
\mathbf{v}_{\text{COM}}
= \mathbf{v}_{\text{marker}}
+ \boldsymbol{\omega} \times \mathbf{r}_{\text{CP}}
\end{equation}
where \(\mathbf{r}_{\text{CP}} = [0.2,\, 0,\, {-0.63}]\)~m is the vector from the marker origin to the COM in the vehicle frame. The corrected COM velocity and the tracked altitude (used to compute instantaneous water level) serve as the kinematic inputs to the surrogate during validation.

\subsection{Validation Pipeline}
Because the vehicle's drivetrain, tire-ground contact, and gravitational forces act simultaneously during each trial, the hydrodynamic force cannot be isolated from sensor measurements alone. Validation was therefore done indirectly. Rather than comparing predicted forces against a measured ground truth, the surrogate's predictions were tested to satisfy a set of well-established hydrodynamic scaling laws.

\subsubsection{Inference and Data Extraction}\label{sec:val_inference}
For each trial, the trained network receives the instantaneous body-frame velocity $(v_x, v_y, v_z)$ and a per-timestep water level $z_w(t)$ computed from the motion-capture altitude and the known water depth, with a pitch-based correction for the marker-to-body-origin offset. The network evaluates all $K{=}10$ active body surfaces at each timestep, producing total predicted hydrodynamic force components
$F_x^{\mathrm{pred}}$ (drag, opposing motion) and $F_z^{\mathrm{pred}}$ (vertical, combining buoyancy and lift).
\par
Metrics are computed exclusively on the \emph{planar section} of each
trial, the segment where the vehicle is traveling on the flat ground floor at approximately constant submersion depth. In both cases, the vehicle traverses the same flat ground floor at the same nominal depth, so the two directions are physically equivalent for the purpose of steady-state hydrodynamic loading. A planar section is accepted when its duration exceeds \SI{0.2}{\second} ($>$20 samples) and its mean vertical velocity satisfies $|\bar{v}_z| < \SI{0.05}{\meter\per\second}$, ensuring approximately constant submersion depth throughout the segment. Of the 65 total trials (39~ramp-in, 26~ramp-out), 57 meet these criteria (36~ramp-in, 21~ramp-out), spanning three depths and body-frame speeds from \SIrange{0.52}{3.78}{\meter\per\second}.

\subsubsection{Physics-Based Test Definitions}\label{sec:val_tests}
Two tests are applied, each probing a distinct aspect of the predicted
forces:
\vspace{3pt}
\paragraph{Test~1: Drag-speed scaling}
At constant depth on a flat surface, hydrodynamic drag on a bluff body
follows;
\begin{equation}\label{eq:drag_v2}
  |F_x^{\mathrm{pred}}| \;=\; C_{D,\mathrm{eff}}\, v^2
\end{equation}
where the effective drag parameter $C_{D,\mathrm{eff}} = \tfrac{1}{2}\rho\, C_D\, A_{\mathrm{proj}}$ lumps the drag coefficient, fluid density, and projected frontal area into a single constant for each depth. The depth used in evaluation is the nominal quiescent water level measured before each set of trials. Although the local free surface is displaced by the bow wave and wake during motion, the surrogate was trained on CFD simulations in which the water surface is defined as the undisturbed far-field level, so the static measurement is the appropriate input. For each trial, the mean drag magnitude and mean squared speed in the planar section yield one data point and an origin-constrained linear fit is performed per depth. The physical drag coefficient is recovered using;
\begin{equation}\label{eq:cd_physical}
  C_D \;=\; \frac{C_{D,\mathrm{eff}}}{\tfrac{1}{2}\,\rho\, W\, d}
\end{equation}
where the projected frontal area is approximated as
$A_{\mathrm{proj}} = W \times d$ (vehicle width $\times$ submersion depth).
As a self-consistency check, speed-matched trial pairs are formed across depths and the ratio $C_{D,\mathrm{eff}}^{(d_2)} / C_{D,\mathrm{eff}}^{(d_1)}$ is compared against the ratio implied by the per-depth aggregate fits. This ensures that the quadratic model holds at the individual trial level.
\vspace{3pt}
\paragraph{Test~2: Vertical force consistency}
The predicted vertical force $F_z^{\mathrm{pred}}$ is decomposed at each
depth via
\begin{equation}\label{eq:fz_decomp}
  F_z \;=\; F_0(d) \;+\; C_L\, v^2
\end{equation}
where $F_0$ is the speed-independent buoyancy intercept and $C_L v^2$
captures dynamic lift. At zero forward speed, the only vertical force
on a partially submerged rigid body is hydrostatic buoyancy
($\rho g V_{\mathrm{sub}}$), so $F_0$ corresponds to the Archimedean
displaced-volume force at each depth.  Any additional vertical load at
$v > 0$ arises from hydrodynamic pressure on the body, which scales
with $v^2$ in the standard dynamic-pressure form. Three sub-checks are applied;
(a)~mean $F_z$ increases monotonically with depth,
(b)~buoyancy intercepts $F_0$ scale linearly with depth, and
(c)~at matched speed bins, deeper submersion produces higher $F_z$.

\section{Results}
This section evaluates the proposed hydrodynamic surrogate in two stages. First, inference metrics and force prediction accuracy are assessed on a held-out CFD test split across both vehicle platforms (Husky, $K{=}13$ surfaces; Warthog, $K{=}10$ surfaces). Second, the trained model is deployed on motion capture data from 57 trials of the Warthog and tested against established hydrodynamic scaling laws. All surrogate evaluations use the same pre-processing and normalizations used during training. Latency and accuracy benchmarks were performed on an Intel i7-13700HX CPU.

\subsection{Inference latency}
Low latency inference is required for closed-loop simulation and onboard deployment. Latency was measured end to end and includes global feature construction, signed-distance submergence evaluation, per-surface feature assembly, normalization, the network forward pass, and force reconstruction.

\begin{figure}
\centering
\includegraphics[width=0.70\linewidth]{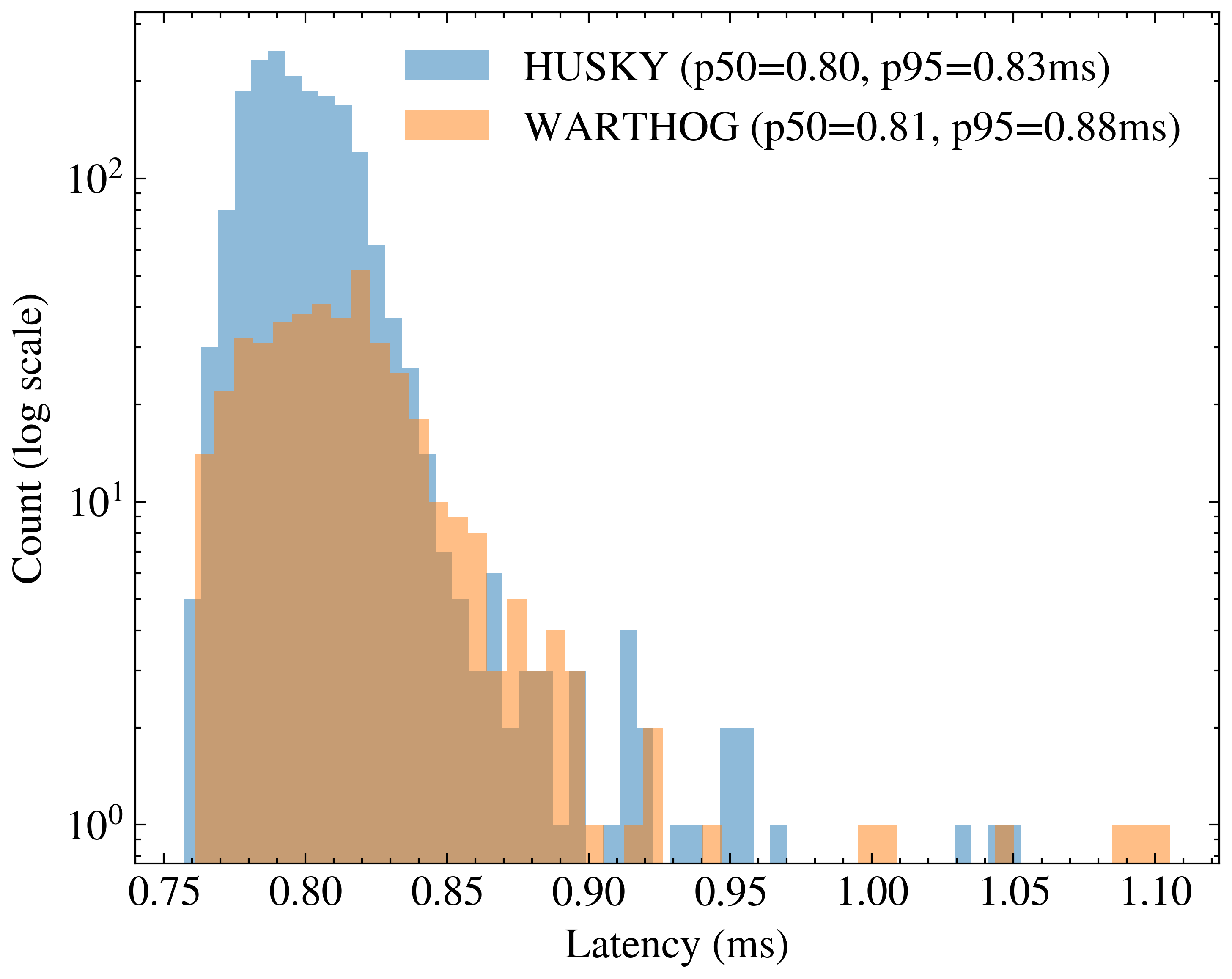}
\caption{End-to-end single-sample CPU latency distribution for Husky and
 Warthog. The count axis is log-scaled to expose tail behaviour.}
\label{fig:latency_hist}
\end{figure}

Table~\ref{tab:latency_summary} reports single-sample CPU latency. The
median latency was \SI{0.826}{\milli\second} on Husky and \SI{0.833}{\milli\second} on Warthog, with $p_{95}$ below \SI{0.90}{\milli\second} on both vehicles. Fig.~\ref{fig:latency_hist} shows the full distribution on a log-scaled count axis.

A sustained inference loop over \SI{300}{\second} on Husky confirmed runtime stability, exceeding \SI{1}{\kilo\hertz} for both constant-velocity and time-varying inputs. The $p_{99}$ latency remained below
\SI{1.6}{\milli\second}, providing margin for real-time integration at
common control rates.

\begin{table}
\centering
\caption{Single-sample end-to-end inference latency on CPU.}
\label{tab:latency_summary}
\begin{tabular}{lrrrrrr}
\hline
Vehicle & $N_{\text{test}}$ & Mean (ms) & Std (ms) & Median (ms) & $p_{95}$ (ms) \\
\hline
Husky   & 1836 & 0.832 & 0.033 & 0.826 & 0.875 \\
Warthog &  433 & 0.838 & 0.033 & 0.833 & 0.898 \\
\hline
\end{tabular}
\end{table}

\subsection{Net Force Accuracy}\label{sec:res_net_force}
Accuracy was evaluated at two levels; net force in Newtons, which directly impacts vehicle dynamics, and per-surface force, which measures local loading fidelity. Because the Warthog (1.52\,m, 280\,kg) operates at higher speeds and deeper submersion than the Husky (0.86\,m, 50\,kg), its hydrodynamic forces are larger in comparison. MAE and RMSE therefore reflect this scale difference rather than model quality. To enable cross-platform comparison, Table~\ref{tab:net_force_accuracy} also reports symmetric MAPE (sMAPE) with a 1\,N floor, which normalizes the error by force magnitude and allows direct comparison of prediction quality across vehicles of different scale.
\par
\begin{figure}
\centering
\includegraphics[width=1\linewidth]{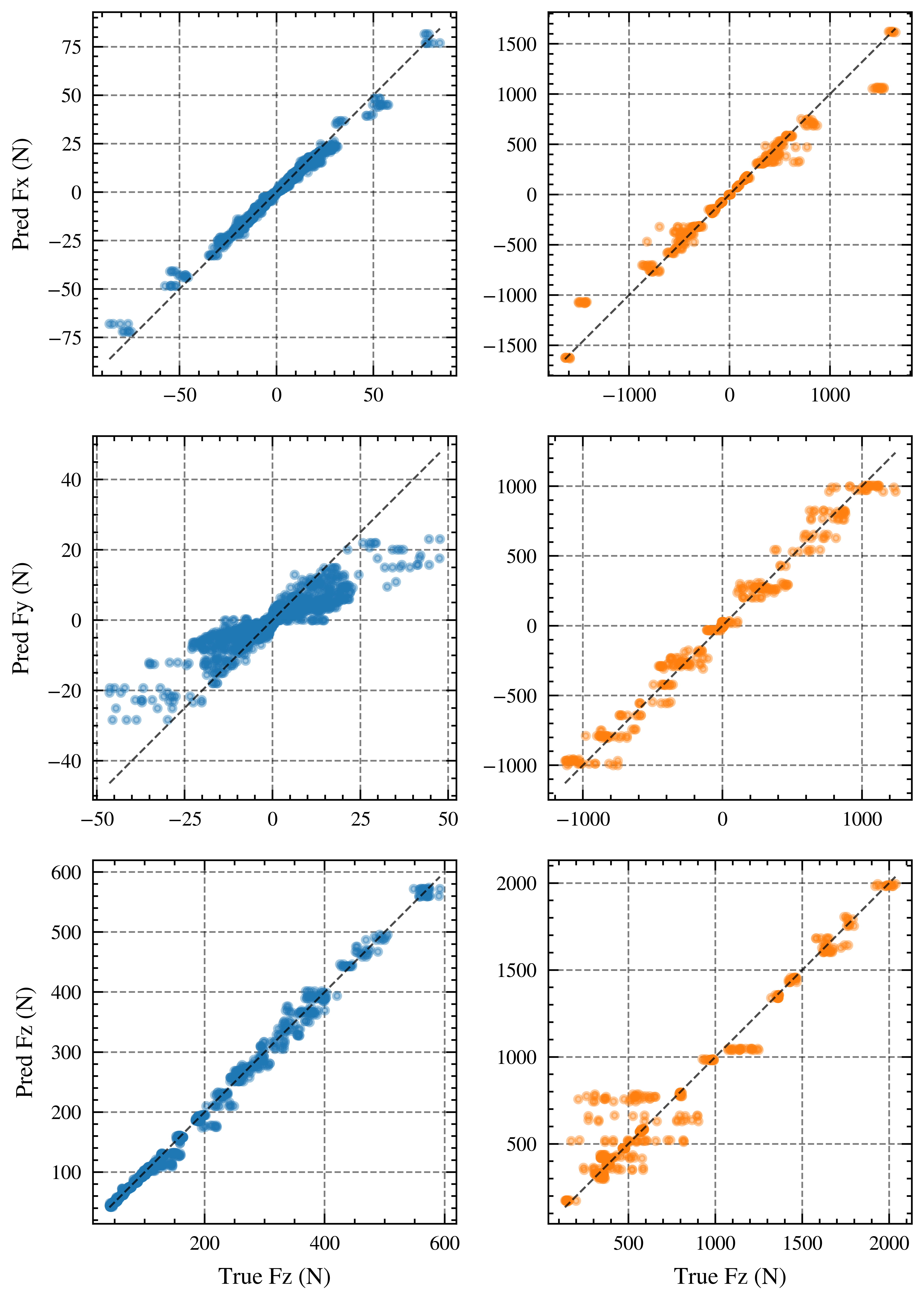}
\caption{Parity plots for net forces. Rows correspond to $F_x$, $F_y$,
 and $F_z$; columns to Husky (left) and Warthog (right). The dashed line is the 1:1 reference.}
\label{fig:parity_net}
\end{figure}

Table~\ref{tab:net_force_accuracy} reports the held-out test metrics.
Fig.~\ref{fig:parity_net} provides corresponding parity plots. The
longitudinal component $F_x$ achieved an sMAPE of approximately 13\% on both vehicles, indicating that the normalized physics and geometry features transfer well across scale. The vertical component $F_z$ is predicted with 2.99\% sMAPE on Husky and 11.77\% on Warthog. The higher Warthog error is consistent with the larger force magnitudes and fewer training samples available for that platform. Lateral force $F_y$ exhibits elevated sMAPE (${\sim}55$\%) for both vehicles. This is expected for a distribution that concentrates near zero and includes frequent sign changes, where relative error measures become sensitive independently of the absolute residual magnitude.
\par
Per-surface MAE was 0.17\,N ($F_x$), 0.40\,N ($F_y$), and 0.63\,N ($F_z$) on Husky, and 8.04\,N, 9.82\,N, and 13.02\,N on Warthog respectively. The net force MAE exceeds the per-surface values because residuals accumulate across the $K$ surfaces in each summation.
\begin{table}
\centering
\caption{Net force accuracy on the held-out test split. Symmetric MAPE uses an $\epsilon{=}1$\,N floor.}
\label{tab:net_force_accuracy}
\begin{tabular}{llrrrr}
\hline
Vehicle & Comp. & MAE (N) & RMSE (N) & sMAPE (\%) \\
\hline
Husky   & $F_x$ &   1.25 &   2.10 & 12.96 \\
Husky   & $F_y$ &   3.42 &   5.64 & 56.19 \\
Husky   & $F_z$ &   4.38 &   7.75 &  2.99 \\
Warthog & $F_x$ &  52.87 & 112.86 & 12.98 \\
Warthog & $F_y$ &  53.28 &  76.70 & 54.84 \\
Warthog & $F_z$ &  67.61 & 117.31 & 11.77 \\
\hline
\end{tabular}
\end{table}

\subsection{Per-Surface Accuracy and Boundary-Region Performance}\label{sec:res_decomp}
Because the network predicts forces per surface, prediction accuracy can be decomposed by geometric category. Wheels account for 79.1\% of the total absolute force on Husky and 82.2\% on Warthog, consistent with the physical expectation that wheel submergence governs a large fraction of hydrodynamic loading on wheeled ground vehicles. On Husky, the largest per-category MAE arises from the wheel surfaces (1.02\,N in $F_x$, 1.60\,N in $F_z$), with the bottom surface contributing 2.99\,N in $F_z$. On Warthog, the bottom category dominates vertical MAE (58.89\,N in $F_z$), while wheels contribute 17.88\,N.
\par
To assess performance at the boundaries of training coverage, accuracy was evaluated on 5th and 95th-percentile slices defined by submersion depth, speed, and oblique flow ratio. On Husky, the deep-submersion slice raises $F_z$ MAE to 8.67\,N and the low-speed slice to 10.27\,N, reflecting increased sensitivity where small changes in depth alter the submergence configuration. On Warthog, the high-speed slice reports $F_x$ MAE of 66.31\,N and $F_z$ MAE of 66.91\,N.

\subsection{Ablation Studies}%
\label{sec:res_ablation}
Three ablation variants were trained on the combined dataset with identical hyperparameters and splits to probe the sensitivity of the architecture to its input features, surface tessellation, and per-surface decomposition.
\par
The first variant removed the explicit vehicle dimensions ($L$, $W$, $H$) from the global feature vector ($D_g{:}\;12{\to}9$). This raised longitudinal-force MAE by 65\% on Husky (1.25$\to$2.06\,N) and 31\% on Warthog (52.87$\to$69.22\,N). Because the Froude and Reynolds numbers still encode a vehicle-specific $L_\text{ref}$, scale information was not entirely removed. The degradation indicates that explicit dimensions provide non-redundant geometric context beyond what the dimensionless groups capture.
\par
The second variant coarsened the surface tessellation by merging the three depth-stratified front and rear sub-patches on Husky into single vertical plates ($K{:}\;13{\to}9$). As a result, net force MAE was preserved but per-surface lateral-force MAE increased by 35\% (0.40$\to$0.54\,N). The merged representation replaces three independent submergence fractions with a single area-weighted average, removing the model's ability to distinguish depth-dependent loading across the front face.
\par
Finally, a global-only MLP was tested that mapped the same global features directly to net $F/\rho$. The resulting model recovers net force MAE within 1-2\,N of the per-surface model on both vehicles (Husky $F_x$: 0.90\,N vs.\ 1.25\,N). The per-surface architecture therefore does not improve net force accuracy over this simpler baseline. However, its value lies in providing spatially resolved forces at each patch (0.17-0.63\,N MAE on Husky, 8-13\,N on Warthog), enabling distributed surface loads for dynamics simulation and real-world behavioral comprehension, capabilities that a single aggregate output cannot provide.

\subsection{Experimental Validation}\label{sec:res_exp_val}
The two physics-based tests described in Sec.~\ref{sec:val_tests}
were applied to the 57 qualifying planar sections (36 ramp-in, 21 ramp-out) extracted from water wading trials of the Warthog across three submersion depths.

\subsubsection{Drag-Speed Scaling}\label{sec:res_drag}
\begin{figure}[t]
  \centering
  \includegraphics[width=0.8\linewidth]{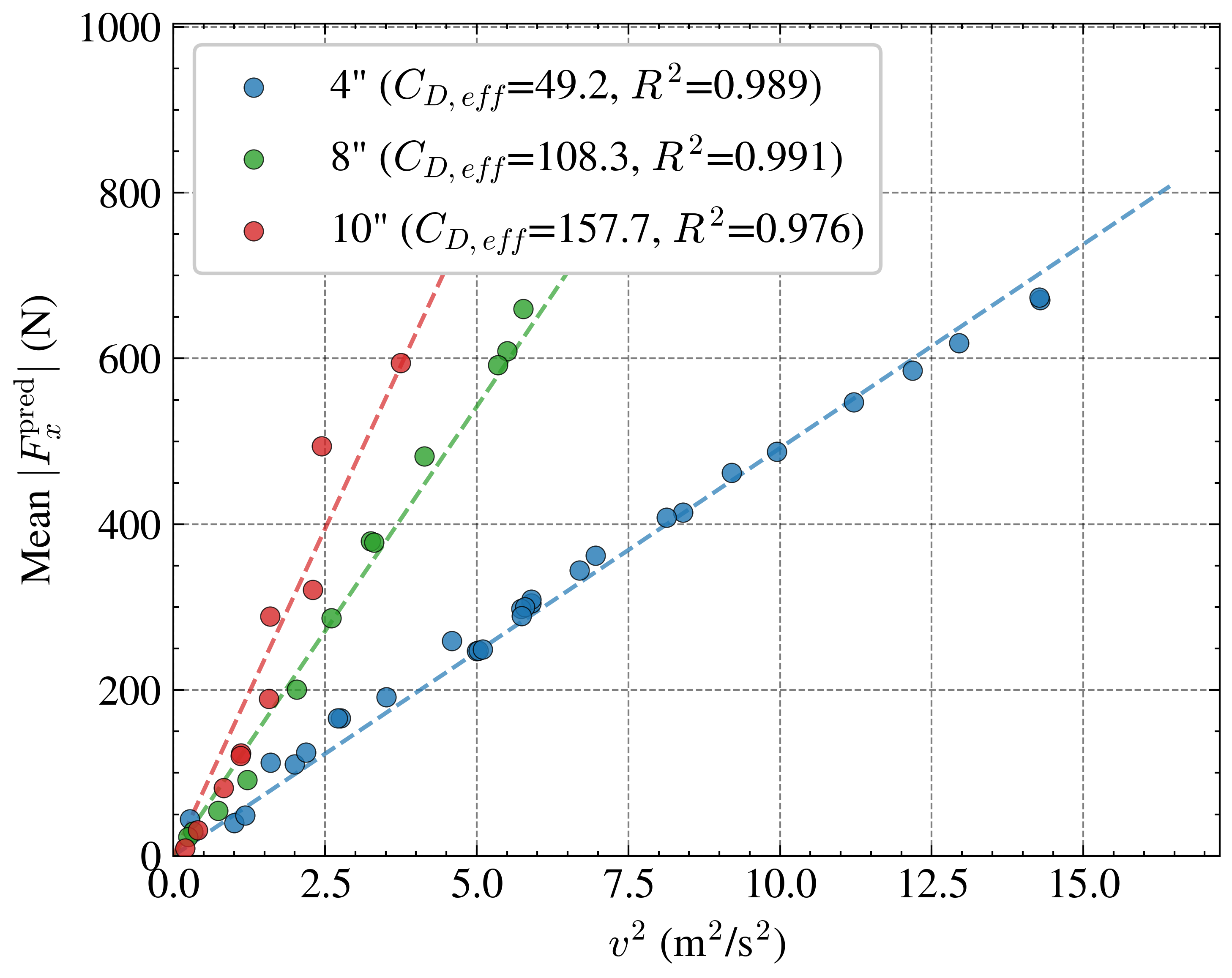}
  \caption{%
    Mean predicted drag $|F_x^{\mathrm{pred}}|$ versus $v^2$
    in the planar ground section at each depth. Dashed lines are per-depth origin-constrained fits ($|F_x| = C_{D,\mathrm{eff}}\, v^2$). Each marker represents one trial.}
  \label{fig:drag_scaling}
\end{figure}

Fig.~\ref{fig:drag_scaling} shows the mean predicted drag magnitude
against the mean squared speed for each trial. The origin-constrained
quadratic fits (Eq.~\ref{eq:drag_v2}) yield;
$R^2 = 0.995$ (\SI{4}{in}, $n{=}29$),
$R^2 = 0.991$ (\SI{8}{in}, $n{=}14$), and
$R^2 = 0.980$ (\SI{10}{in}, $n{=}14$).
The corresponding effective drag parameters are
$C_{D,\mathrm{eff}} = 49.2$, $108.3$, and
$\SI{157.7}{\newton\second\squared\per\meter\squared}$.
\par
At \SI{4}{in} depth, the 29 data points span speeds from \SI{0.52}{} to
\SI{3.8}{\meter\per\second}. The quadratic relationship $|F_x| \propto v^2$ was not encoded in the training objective, yet the network reproduced it across the full tested speed range. Extracting the physical drag coefficient using Eq.~\eqref{eq:cd_physical} yields $C_D = 0.70$ (\SI{4}{in}), $0.77$ (\SI{8}{in}), and $0.90$ (\SI{10}{in}). The monotonic increase with depth is consistent with the vehicle geometry. At greater submersion the wheel assemblies and chassis underside become exposed to the flow, increasing the effective block. The $C_{D,\mathrm{eff}}$ values also grow monotonically with submersion, following the near-linear growth of projected frontal area $A_{\mathrm{proj}} = W d$ modulated by the gradual increase in $C_D$ described above.

To verify that this scaling was internally consistent at the individual
trial level, speed-matched trial pairs were formed across all three depth combinations (tolerance $\pm$\SI{0.3}{\meter\per\second}, minimum speed \SI{0.6}{\meter\per\second}) and the ratio of per-trial $C_{D,\mathrm{eff}}^{i} = |F_x^{i}|/v_i^2$ values was computed for
each pair. Across 177 matched pairs, the empirical mean ratios agree with the ratios implied by the per-depth aggregate fits to within 1-9\%,
confirming that the quadratic drag model generalizes at the individual
trial level and is not an artifact of the aggregate fitting procedure.

\subsubsection{Vertical Force Consistency}\label{sec:res_fz}
The vertical force predictions provide an independent line of evidence that probes a different physical mechanism i.e., hydrostatic buoyancy rather than dynamic drag. Mean $F_z^{\mathrm{pred}}$ increased monotonically with depth; \SI{442}{\newton} (\SI{4}{in}), \SI{650}{\newton} (\SI{8}{in}), and \SI{816}{\newton} (\SI{10}{in}), as expected from the increasing displaced volume.
\par
\begin{figure}[t]
  \centering
  \includegraphics[width=0.8\linewidth]{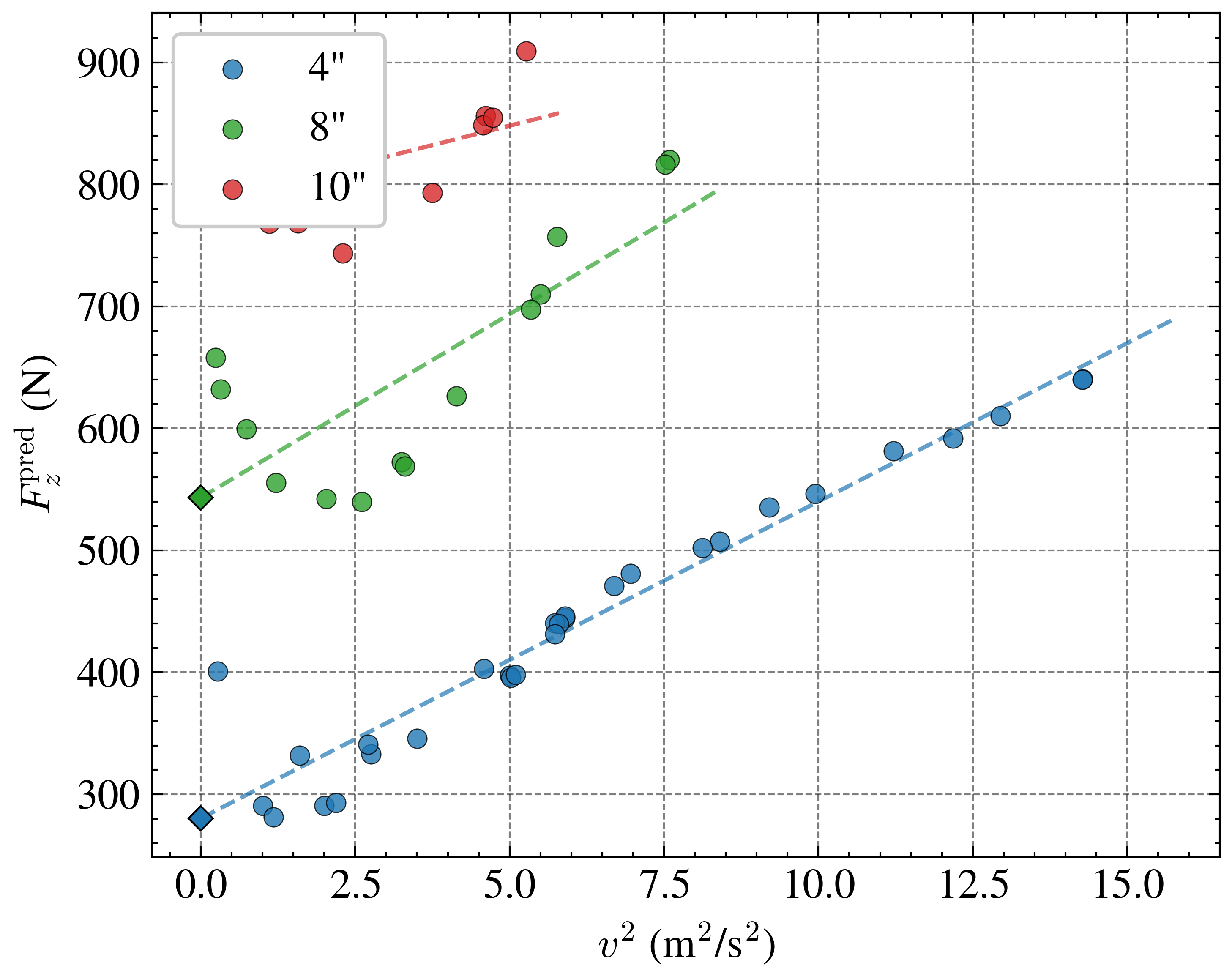}
  \caption{%
    Predicted vertical force $F_z$ versus $v^2$ at each depth with
    per-depth linear fits ($F_z = F_0 + C_L\, v^2$). Diamonds mark
    the speed-independent buoyancy intercepts $F_0$.}
  \label{fig:fz_consistency}
\end{figure}

Decomposing $F_z$ into a buoyancy intercept $F_0$ and a dynamic lift component $C_L v^2$ (Eq.~\ref{eq:fz_decomp}) yields speed-independent buoyancy terms of \SI{280}{\newton}, \SI{543}{\newton}, and \SI{784}{\newton}, as shown in Fig. ~\ref{fig:fz_consistency}. These intercepts scale linearly with depth ($R^2 = 0.973$), consistent with hydrostatic buoyancy acting on a body of approximately constant horizontal cross-section. As with the drag test, the network training loss contains no explicit buoyancy or hydrostatic pressure term; the linear $F_0$-depth relationship is a consequence of the per-surface architecture correctly integrating pressure contributions from each submerged patch. At fixed speed bins, deeper submersion consistently produces larger $F_z$, confirming that the monotonic trend is not an artifact of different speed distributions across depth conditions.
\par
The ratio $F_0^{10''}/F_0^{4''} \approx 2.8$ exceeds the depth ratio $d_{10}/d_4 = 2.5$, which is consistent with the non-prismatic hull geometry: at greater submersion additional volume from the wheel wells and chassis underside contributes to the displaced fluid, producing a buoyancy increment beyond what a simple rectangular cross-section predicts. This mirrors the depth-dependent $C_D$ increase observed in the drag test and provides independent confirmation that the surrogate
resolves geometry-specific submersion effects. The dynamic lift component remains small relative to buoyancy across the tested speed range, consistent with the moderate Froude numbers ($Fr_h < 3$) of the experiments. At \SI{10}{in} depth, buoyancy accounts for over 95\% of the total vertical force, leaving the speed-dependent lift term difficult to resolve from the inter-trial scatter; the per-depth linear fit accordingly explains less variance at greater submersion ($R^2 = 0.93$ at \SI{4}{in} versus $0.23$ at \SI{10}{in}). The primary validation therefore rests on the buoyancy intercepts $F_0$ and their linear scaling with depth, which are robust to this effect.

\subsubsection{Sim-to-Real Transfer}\label{sec:res_sim2real}
Taken together, the drag-scaling and vertical-force tests constitute direct evidence of physics-consistent sim-to-real transfer. The surrogate was trained exclusively on CFD data with no real-world parameter tuning yet when driven by measured kinematics from experimental trials, its predictions satisfy two independent physical scaling laws-quadratic drag ($R^2 \geq 0.97$) and linear buoyancy-depth ($R^2 = 0.973$); neither of which is encoded in the training loss. Both emerge from the per-surface architecture summing individually predicted patch forces. The depth-dependent increase in $C_D$ and the buoyancy intercept ratio exceeding the depth ratio further demonstrate that the transfer preserves geometry-specific physical effects. These results demonstrate that the surrogate's learned representation transfers from simulation to real-world operating conditions while preserving the geometry-dependent physical structure of the hydrodynamic loading.

\subsection{Limitations}
\paragraph{Quasi-steady training regime}
Although each CFD simulation captures the transient evolution of the free surface, the 20 snapshots sampled per run are treated as independent samples rather than a time series. The surrogate therefore learns an instantaneous state-to-force mapping and cannot represent history-dependent phenomena such as added-mass transients or vortex-shedding periodicity. The ordered snapshot sequences already present in the dataset could support a frame-buffer extension without additional CFD campaigns.
\paragraph{Indirect force validation}
On a wheeled ground vehicle in water, drivetrain torque, tire-ground reaction, fluid forces, and gravity all act on the vehicle simultaneously and no onboard sensor can isolate the hydrodynamic component from the total load. This is an inherent instrumentation constraint of any driven ground vehicle in water. Experimental validation therefore evaluates the surrogate's predicted forces, which are purely hydrodynamic by construction, by checking whether they reproduce expected physical behavior i.e., quadratic drag scaling with speed and linear buoyancy scaling with depth.
\paragraph{Geometric fidelity}
The vehicle geometry is represented by $K$ flat surface patches ($K{=}10$ for Warthog, $K{=}13$ for Husky), approximating curved and articulated components such as wheels and suspension. The SDF-based submergence calculation further assumes a rigid body, so deformation under load is not captured. The ablation study shows that coarser tessellation preserves net force accuracy but degrades per-surface resolution, suggesting that adaptive patch subdivision in regions of high geometric curvature could be further explored.

\section{Conclusion \& Future Work}
This work presented a per-surface neural surrogate for hydrodynamic force prediction on ground vehicles, trained on high-fidelity CFD data from two geometrically distinct platforms. By decomposing forces at the surface level, the model captures how individual body regions contribute to the net load as a function of depth and flow. Experimental validation on real-world trials showed that the
CFD-trained surrogate reproduces quadratic drag scaling and linear buoyancy-depth relationships that are absent from the training objectives.
\par
Three aspects of the results merit emphasis. First, the near-identical longitudinal-force sMAPE across Husky and Warthog (13\%) demonstrates that the normalized geometry and physics features enable cross-platform generalization without vehicle-specific retraining. Second, the depth-dependent variation in $C_D$ and $F_0$ is not a property that a lumped-parameter model, one that maps a single depth value to a single force scalar, can reproduce. It arises because the per-surface decomposition resolves which body panels are submerged at each depth and how each contributes to the net load. Third, the surrogate was trained entirely in simulation yet reproduces established hydrodynamic scaling laws on real-world data, providing direct evidence of physics-consistent sim-to-real transfer.
\par
The framework offers a pathway towards real-time, geometry-resolved hydrodynamic modeling for field robotics. Sub-millisecond CPU inference and cross-platform generalization across two vehicles of different size and geometry make the surrogate a candidate force module for onboard controllers and simulation environments. Future work includes extending the single-frame input to a temporal buffer for history-dependent dynamics, integrating the surrogate into a real-time Unreal Engine 5 simulation loop for closed-loop validation, and applying the framework to additional vehicle platforms to further test and validate cross-geometry generalization.

\section*{Acknowledgments}
The authors would like to thank the Texas A\&M High Performance Research Computing for the advanced computing resources; Mohini Priya Kolluri for the invaluable contributions towards the initial CFD framework and experiments; Morgan Jenks for assistance with the motion capture system.

This document is an overview of UK MOD’s Defence Science and Technology Laboratory (DSTL) sponsored research and is released for informational purposes only. The contents of this document should not be interpreted as representing the views of the UK MOD, nor should it be assumed that they reflect any current or future UK MOD policy.

The authors further disclose that AI tools were used to improve the readability of the text, and they accept full responsibility for the content of the final publication.

\bibliographystyle{IEEEtran}
\bibliography{References}

@article{waheed2025quantifying,
  title={Quantifying the Sim2Real Gap: Model-Based Verification and Validation in Autonomous Ground Systems},
  author={Waheed, Ammar and Areti, Madhu and Gallantree, Luke and Hasnain, Zohaib},
  journal={IEEE Robotics and Automation Letters},
  year={2025},
  publisher={IEEE}
}

@article{de2022simu2vita,
  title={Simu2vita: A general purpose underwater vehicle simulator},
  author={de Cerqueira Gava, Pedro Daniel and Nascimento J{\'u}nior, Cairo L{\'u}cio and Belchior de Fran{\c{c}}a Silva, Juan Ram{\'o}n and Adabo, Geraldo Jos{\'e}},
  journal={Sensors},
  volume={22},
  number={9},
  pages={3255},
  year={2022},
  publisher={MDPI}
}

@article{wang2023prediction,
  title={The prediction of external flow field and hydrodynamic force with limited data using deep neural network},
  author={Wang, Tong-sheng and Xi, Guang and Sun, Zhong-guo and Huang, Zhu},
  journal={Journal of Hydrodynamics},
  volume={35},
  number={3},
  pages={549--570},
  year={2023},
  publisher={Springer}
}

@article{bai2022data,
  title={Data-driven prediction of experimental hydrodynamic data of the manta ray robot using deep learning method},
  author={Bai, Jingyi and Huang, Qiaogao and Pan, Guang and He, Junjie},
  journal={Journal of Marine Science and Engineering},
  volume={10},
  number={9},
  pages={1285},
  year={2022},
  publisher={MDPI}
}

@article{seyed2022physics,
  title={Physics-inspired architecture for neural network modeling of forces and torques in particle-laden flows},
  author={Seyed-Ahmadi, Arman and Wachs, Anthony},
  journal={Computers \& Fluids},
  volume={238},
  pages={105379},
  year={2022},
  publisher={Elsevier}
}

@INPROCEEDINGS{9982036,
  author={Angelidis, Emmanouil and Bender, Jan and Arreguit, Jonathan and Gleim, Lars and Wang, Wei and Axenie, Cristian and Knoll, Alois and Ijspeert, Auke},
  booktitle={2022 IEEE/RSJ International Conference on Intelligent Robots and Systems (IROS)}, 
  title={Gazebo Fluids: SPH-based simulation of fluid interaction with articulated rigid body dynamics}, 
  year={2022},
  volume={},
  number={},
  pages={11238-11245},
  doi={10.1109/IROS47612.2022.9982036}}

@INPROCEEDINGS{UNav,
  author={Amer, Abdelhakim and {\'A}lvarez-Tu{\~n}{\'o}n, Olaya and U{\u{g}}urlu, Halil {\.I}brahim and Le Fevre Sejersen, Jonas and Brodskiy, Yury and Kayacan, Erdal},
  booktitle={2023 21st International Conference on Advanced Robotics (ICAR)}, 
  title={UNav-Sim: A Visually Realistic Underwater Robotics Simulator and Synthetic Data-Generation Framework}, 
  year={2023},
  volume={},
  number={},
  pages={570-576},
  doi={10.1109/ICAR58858.2023.10406819}}

@article{choi2021use,
  title={On the use of simulation in robotics: Opportunities, challenges, and suggestions for moving forward},
  author={Choi, HeeSun and Crump, Cindy and Duriez, Christian and Elmquist, Asher and Hager, Gregory and Han, David and Hearl, Frank and Hodgins, Jessica and Jain, Abhinandan and Leve, Frederick and others},
  journal={Proceedings of the National Academy of Sciences},
  volume={118},
  number={1},
  pages={e1907856118},
  year={2021},
  publisher={National Academy of Sciences}
}

@INPROCEEDINGS{DAVE,
  author={Zhang, Mabel M. and Choi, Woen-Sug and Herman, Jessica and Davis, Duane and Vogt, Carson and McCarrin, Michael and Vijay, Yadunund and Dutia, Dharini and Lew, William and Peters, Steven and Bingham, Brian},
  booktitle={2022 IEEE/OES Autonomous Underwater Vehicles Symposium (AUV)}, 
  title={DAVE Aquatic Virtual Environment: Toward a General Underwater Robotics Simulator}, 
  year={2022},
  volume={},
  number={},
  pages={1-8},
  doi={10.1109/AUV53081.2022.9965808}}

@INPROCEEDINGS{Stonefish,
  author={Cieślak, Patryk},
  booktitle={OCEANS 2019 - Marseille}, 
  title={Stonefish: An Advanced Open-Source Simulation Tool Designed for Marine Robotics, With a ROS Interface}, 
  year={2019},
  volume={},
  number={},
  pages={1-6},
  doi={10.1109/OCEANSE.2019.8867434}}

@misc{song2025oceansim,
      title={OceanSim: A GPU-Accelerated Underwater Robot Perception Simulation Framework}, 
      author={Jingyu Song and Haoyu Ma and Onur Bagoren and Advaith V. Sethuraman and Yiting Zhang and Katherine A. Skinner},
      year={2025},
      eprint={2503.01074},
      archivePrefix={arXiv},
      primaryClass={cs.RO},
      url={https://arxiv.org/abs/2503.01074}, 
}

@InProceedings{pmlr-v78-dosovitskiy17a,
  title = 	 {{CARLA}: {An} Open Urban Driving Simulator},
  author = 	 {Dosovitskiy, Alexey and Ros, German and Codevilla, Felipe and Lopez, Antonio and Koltun, Vladlen},
  booktitle = 	 {Proceedings of the 1st Annual Conference on Robot Learning},
  pages = 	 {1--16},
  year = 	 {2017},
  editor = 	 {Levine, Sergey and Vanhoucke, Vincent and Goldberg, Ken},
  volume = 	 {78},
  series = 	 {Proceedings of Machine Learning Research},
  month = 	 {13--15 Nov},
  publisher =    {PMLR},
  url = 	 {https://proceedings.mlr.press/v78/dosovitskiy17a.html}
}

@INPROCEEDINGS{s2r-survey,
  author={Zhao, Wenshuai and Queralta, Jorge Peña and Westerlund, Tomi},
  booktitle={2020 IEEE Symposium Series on Computational Intelligence (SSCI)}, 
  title={Sim-to-Real Transfer in Deep Reinforcement Learning for Robotics: a Survey}, 
  year={2020},
  volume={},
  number={},
  pages={737-744},
  doi={10.1109/SSCI47803.2020.9308468}}

@article{GU2018343,
title = {Drag, added mass and radiation damping of oscillating vertical cylindrical bodies in heave and surge in still water},
journal = {Journal of Fluids and Structures},
volume = {82},
pages = {343-356},
year = {2018},
issn = {0889-9746},
doi = {https://doi.org/10.1016/j.jfluidstructs.2018.06.012},
url = {https://www.sciencedirect.com/science/article/pii/S0889974617306552},
author = {Hanbin Gu and Peter Stansby and Tim Stallard and Efrain {Carpintero Moreno}}
}

@article{matveev2025,
  title={Hydrodynamics of Semi-Submersible Vehicle Hulls With Variable Height--Width Ratio in Deep and Shallow Water},
  author={Matveev, Konstantin I},
  journal={Journal of Offshore Mechanics and Arctic Engineering},
  volume={147},
  number={6},
  pages={061402},
  year={2025},
  publisher={American Society of Mechanical Engineers}
}

@article{PAN2023114618,
title = {A review on drag reduction technology: Focusing on amphibious vehicles},
journal = {Ocean Engineering},
volume = {280},
pages = {114618},
year = {2023},
issn = {0029-8018},
doi = {https://doi.org/10.1016/j.oceaneng.2023.114618},
url = {https://www.sciencedirect.com/science/article/pii/S0029801823010028},
author = {Dibo Pan and Xiaojun Xu and Bolong Liu and Haijun Xu and Xiaocong Wang}
}

@article{kamath2017study,
  title={Study of water impact and entry of a free falling wedge using computational fluid dynamics simulations},
  author={Kamath, Arun and Bihs, Hans and Arntsen, {\O}ivind A},
  journal={Journal of Offshore Mechanics and Arctic Engineering},
  volume={139},
  number={3},
  pages={031802},
  year={2017},
  publisher={American Society of Mechanical Engineers}
}

@article{huang2021cfd,
  title={CFD analyses on the water entry process of a freefall lifeboat},
  author={Huang, Luofeng and Tavakoli, Sasan and Li, Minghao and Dolatshah, Azam and Pena, Blanca and Ding, Boyin and Dashtimanesh, Abbas},
  journal={Ocean engineering},
  volume={232},
  pages={109115},
  year={2021},
  publisher={Elsevier}
}

@book{horne1963phenomena,
  title={Phenomena of pneumatic tire hydroplaning},
  author={Horne, Walter B and Dreher, Robert C},
  volume={2056},
  year={1963},
  publisher={National Aeronautics and Space Administration}
}

@techreport{varshney2022,
  title={Transient, 3D CFD, Moving Mesh Simulation of Vehicle Water Wading in a Water Tunnel with Inclined Entry-Exit},
  author={Varshney, Mehul and Ballani, Abhishek and Pasunurthi, ShyamSundar and Maiti, Dipak and Dhar, Sujan and Ding, Hui},
  year={2022},
  institution={SAE Technical Paper}
}

@article{aldhaheri2025,
  title={Underwater Robotic Simulators Review for Autonomous System Development},
  author={Aldhaheri, Sara and Hu, Yang and Xie, Yongchang and Wu, Peng and Kanoulas, Dimitrios and Liu, Yuanchang},
  journal={arXiv preprint arXiv:2504.06245},
  year={2025}
}

@article{yamashita2024,
  title={Modeling of Vehicle Mobility in Shallow Water With Data-Driven Hydrodynamics Model},
  author={Yamashita, Hiroki and Martin, Juan Ezequiel and Tison, Nathan and Grunin, Arkady and Jayakumar, Paramsothy and Sugiyama, Hiroyuki},
  journal={Journal of computational and nonlinear dynamics},
  volume={19},
  number={7},
  pages={071010},
  year={2024},
  publisher={American Society of Mechanical Engineers}
}

@article{more2014stability,
  title={Stability and drag analysis of wheeled amphibious vehicle using CFD and model testing techniques},
  author={More, RR and Adhav, Piyush and Senthilkumar, K and Trikande, MW},
  journal={Applied Mechanics and Materials},
  volume={592},
  pages={1210--1219},
  year={2014},
  publisher={Trans Tech Publ}
}

@article{bocanegra2020review,
  title={Review and analysis of vehicle stability models during floods and proposal for future improvements},
  author={Bocanegra, Ricardo A and Vall{\'e}s-Mor{\'a}n, Francisco J and Franc{\'e}s, F{\'e}lix},
  journal={Journal of flood risk management},
  volume={13},
  pages={e12551},
  year={2020},
  publisher={Wiley Online Library}
}

@article{xia2014criterion,
  title={Criterion of vehicle stability in floodwaters based on theoretical and experimental studies},
  author={Xia, Junqiang and Falconer, Roger A and Xiao, Xuanwei and Wang, Yejiang},
  journal={Natural hazards},
  volume={70},
  number={2},
  pages={1619--1630},
  year={2014},
  publisher={Springer}
}

@article{hu2023experimental,
  title={Experimental testing to determine stability thresholds for partially submerged vehicles at different flow orientations},
  author={Hu, Xiaozhe and Li, Junqi and Wang, Wenhai and Fang, Xing},
  journal={Journal of Hydrology},
  volume={620},
  pages={129525},
  year={2023},
  publisher={Elsevier}
}

@article{liu2023resistance,
  title={Resistance reduction optimization of an amphibious transport vehicle},
  author={Liu, Bolong and Xu, Xiaojun and Pan, Dibo},
  journal={Ocean Engineering},
  volume={280},
  pages={114854},
  year={2023},
  publisher={Elsevier}
}

@book{fossen2021handbook,
  title={Handbook of marine craft hydrodynamics and motion control},
  author={Fossen, Thor I},
  year={2011},
  publisher={John wiley \& sons}
}

@inproceedings{manhaes2016uuvsim,
  title={UUV simulator: A gazebo-based package for underwater intervention and multi-robot simulation},
  author={Manh{\~a}es, Musa Morena Marcusso and Scherer, Sebastian A and Voss, Martin and Douat, Luiz Ricardo and Rauschenbach, Thomas},
  booktitle={Oceans 2016 Mts/Ieee Monterey},
  pages={1--8},
  year={2016},
  organization={Ieee}
}

@article{macklin2014unified,
  title={Unified particle physics for real-time applications},
  author={Macklin, Miles and M{\"u}ller, Matthias and Chentanez, Nuttapong and Kim, Tae-Yong},
  journal={ACM Transactions on Graphics (TOG)},
  volume={33},
  number={4},
  pages={1--12},
  year={2014},
  publisher={ACM New York, NY, USA}
}

@article{raissi2019physics,
  title={Physics-informed neural networks: A deep learning framework for solving forward and inverse problems involving nonlinear partial differential equations},
  author={Raissi, Maziar and Perdikaris, Paris and Karniadakis, George E},
  journal={Journal of Computational physics},
  volume={378},
  pages={686--707},
  year={2019},
  publisher={Elsevier}
}

@book{newman1977marine,
  title={Marine hydrodynamics},
  author={Newman, John Nicholas},
  year={2018},
  publisher={MIT press}
}

@inproceedings{tasora2015chrono,
  title={Chrono: An open source multi-physics dynamics engine},
  author={Tasora, Alessandro and Serban, Radu and Mazhar, Hammad and Pazouki, Arman and Melanz, Daniel and Fleischmann, Jonathan and Taylor, Michael and Sugiyama, Hiroyuki and Negrut, Dan},
  booktitle={international conference on high performance computing in science and engineering},
  pages={19--49},
  year={2015},
  organization={Springer}
}

@article{hou2012numerical,
  title={Numerical methods for fluid-structure interaction—a review},
  author={Hou, Gene and Wang, Jin and Layton, Anita},
  journal={Communications in Computational Physics},
  volume={12},
  number={2},
  pages={337--377},
  year={2012},
  publisher={Cambridge University Press}
}

@article{hirt1981volume,
  title={Volume of fluid (VOF) method for the dynamics of free boundaries},
  author={Hirt, Cyril W and Nichols, Billy D},
  journal={Journal of computational physics},
  volume={39},
  number={1},
  pages={201--225},
  year={1981},
  publisher={Elsevier}
}

@misc{epic_water_buoyancy_component,
  author       = {{Epic Games}},
  title        = {Water Buoyancy Component in Unreal Engine},
  howpublished = {\url{https://dev.epicgames.com/documentation/en-us/unreal-engine/water-buoyancy-component-in-unreal-engine}},
  note         = {Epic Developer Community, Unreal Engine 5.7 Documentation},
  year         = {2026}
}

\end{document}